\documentclass{article}

\usepackage[T1]{fontenc}
\usepackage{iclr2025_conference,times}

\usepackage{amsmath,amsfonts,bm}
\usepackage{multicol}

\def\eqref#1{equation~\ref{#1}}

\def\1{\bm{1}}

\DeclareMathAlphabet{\mathsfit}{\encodingdefault}{\sfdefault}{m}{sl}
\SetMathAlphabet{\mathsfit}{bold}{\encodingdefault}{\sfdefault}{bx}{n}

\usepackage{amsmath}
\usepackage{amssymb}
\usepackage{array}
\usepackage{booktabs}
\usepackage{float}
\usepackage{graphicx}
\usepackage{multirow}
\usepackage{xcolor}
\usepackage{hyperref}
\usepackage{url}
\newcommand*\samethanks[1][\value{footnote}]{\footnotemark[#1]}

\definecolor{uclablue}{rgb}{0.15, 0.45, 0.68}
\hypersetup{
    breaklinks,
    citecolor=uclablue,
    colorlinks=true,
}

\usepackage[T1]{fontenc}
\usepackage[utf8]{inputenc}
\usepackage[most]{tcolorbox}
\usepackage{caption}
\usepackage{enumitem}
\usepackage{iftex}
\ifPDFTeX
  \usepackage{CJKutf8}
  \newcommand{\zhtext}[1]{\begin{CJK*}{UTF8}{gbsn}#1\end{CJK*}}
\else
  ="Songti SC" at 10pt
  \newcommand{\zhtext}[1]{{\zhfont #1}}
\fi

\makeatletter
\def\pgf@sys@bp@correct#1{#1=0.99627#1}
\makeatother

\newtcolorbox{promptbox}{
  enhanced,
  breakable,
  colback=gray!8,
  colframe=gray!55,
  boxrule=0.6pt,
  arc=3mm,
  outer arc=3mm,
  left=3mm,
  right=3mm,
  top=2.5mm,
  bottom=2.5mm,
  before skip=8pt,
  after skip=3pt,
  fontupper=\fontfamily{ptm}\selectfont\small
}

\newcommand{\PromptCode}[1]{%
  {\fontfamily{pcr}\selectfont #1}%
}

\newcommand{\PromptTag}[1]{%
  \PromptCode{\textless#1\textgreater}%
}

\newcommand{\PromptEndTag}[1]{%
  \PromptCode{\textless/#1\textgreater}%
}

\newcommand{\PromptHeading}[1]{%
  \par\smallskip\noindent\textbf{#1}\par\smallskip
}

\newcommand{\PromptCaption}[2]{%
  \par
  \addvspace{6pt}%
  \begingroup
  \captionsetup{
    font=small,
    labelfont=bf,
    textfont=rm,
    aboveskip=0pt,
    belowskip=6pt
  }%
  \captionof{figure}{#1}%
  \label{#2}%
  \par
  \endgroup
}

\newcommand{\method}{\textsc{CocoonBench}}

\title{\method: Can Multilingual Search Agents Escape Language Information \mbox{Cocoons?}}

\author{
Hongchuan Zeng$^{1,}$\thanks{Equal contribution.},
Bowen Liu$^{1,}$\samethanks,
Changjiang Gao$^{1,2,3}$,
Senyu Han$^{1,4,5}$,
Zhipei Xu$^{1}$ \\[0.3em]
Peisong Wang$^{6}$,
Xinbing Liang$^{1}$,
Yuhan Li$^{1}$,
Nuo Chen$^{1}$,
Blaze Chen$^{1,}$\thanks{Correspondence to: blazeechen@tencent.com}
\\[0.8em]
\parbox{0.8\textwidth}{%
  \centering
  \footnotesize
  $^{1}$Hunyuan Team, Tencent\\[0.2em]
  $^{2}$National Key Laboratory for Novel Software Technology,
  Nanjing University\\[0.2em]
  $^{3}$Department of Linguistics and Translation,
  City University of Hong Kong\\[0.2em]
  $^{4}$X-LANCE Lab, School of Computer Science,
  Shanghai Jiao Tong University\\[0.2em]
  $^{5}$Shanghai Innovation Institution
  \qquad
  $^{6}$HKUST-GZ
}
}

\begin{document}

\maketitle

\begin{abstract}
    Multilingual search agents can access the global Web, yet may remain trapped in a language environment with insufficient evidence; we refer to this failure as a language information cocoon.
    However, existing search benchmarks largely operate within a single-language, especially English, overlooking entity connections across multiple languages.
    We introduce \method{}, a benchmark that evaluates whether multilingual search agents can escape such language information cocoons.
    \method{} mines answer-centered relational paths from a frozen Wikidata-derived graph and selects clue sets that are individually necessary and jointly identify a unique answer.
    Using language-specific entity names, it records each task's minimum language cover, removes direct identifiers, and renders each task as matched prompts in three languages together with its construction certificate and provenance.
    Across 10 state-of-the-art models, mean accuracy on \method{} ranges from 33.8\% (DeepSeek-V4-Flash) to 69.7\% (Claude-Opus-4.8), leaving even the strongest system below 70\%.
    The same construction certificates support path-aligned trajectory analysis, enabling diagnosis of model failures in language routing, evidence-path recovery, and answer use.
    The result shows that escaping a language cocoon requires more than
    switching languages: successful cross-lingual search depends on both
    covering the correct languages, and connecting the evidence found across
    them.
\end{abstract}

\section{Introduction}
\label{sec:introduction}

Web-searching agents based on large language models (LLMs) alternate between reasoning and external
actions such as querying search engines and fetching web pages~\citep{yao2023react}. Since information is unevenly distributed on the multilingual internet, e.g. Wikipedia contents can be substantially different across language versions \citep{bao2012omnipedia,wulczyn2016growing,roy2020information}, it is important for the agents to optimize query languages to maximize searching efficiency. 

This creates a problem for search agents. When the evidence exposed in the current language is insufficient, an agent may continue reformulating queries in the same language, fall back to English, or switch to another language that appears relevant but does not contain the missing evidence. We call this failure to leave an insufficient language environment a \emph{language information cocoon}: the agent remains confined to a language-conditioned view of the Web even when solving the task requires evidence from another language environment.

\begin{figure}[t]
    \centering
    \includegraphics[width=\linewidth]{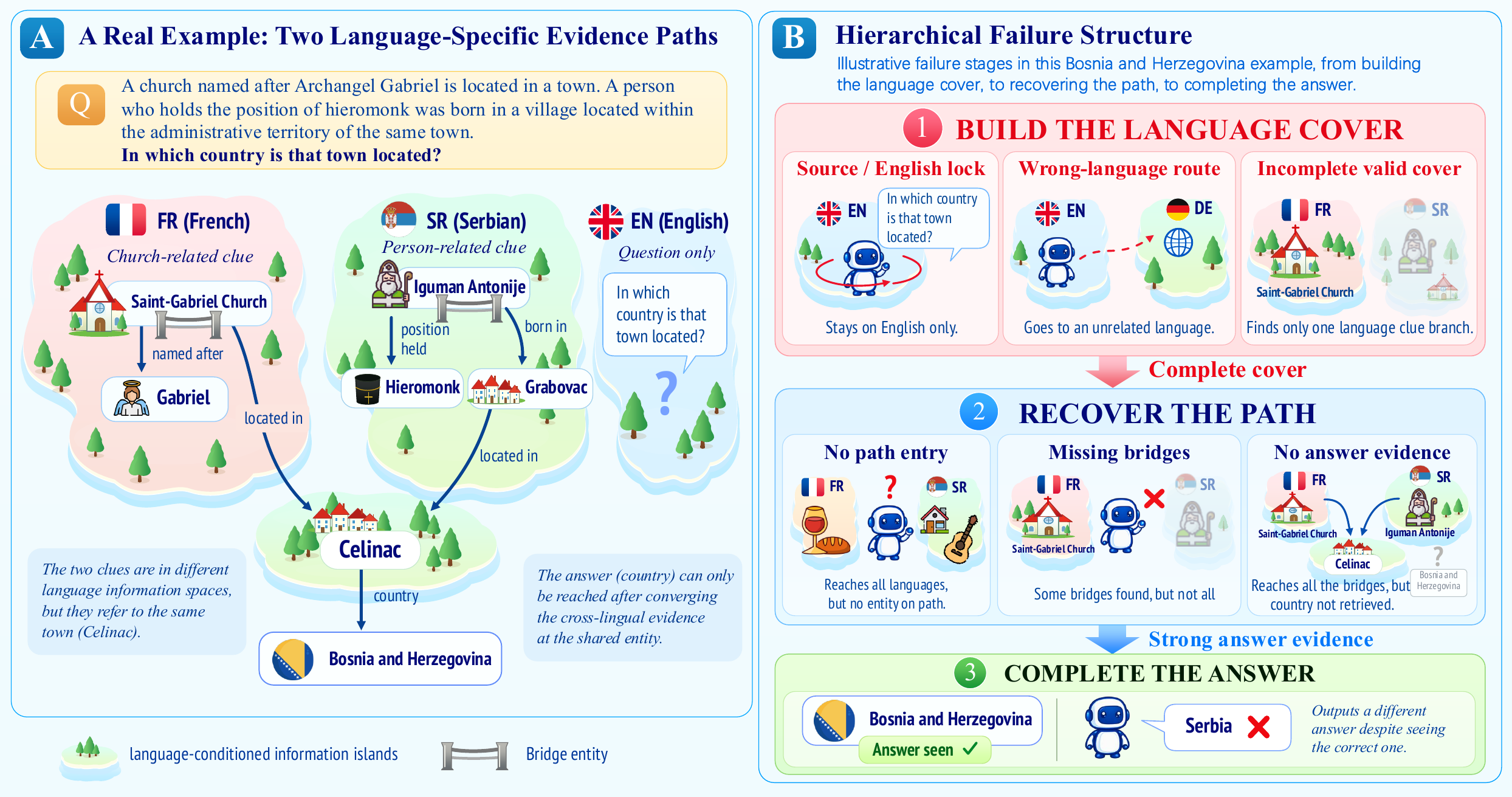}
    \caption{A \method{} task and where agents fail on it. \textbf{(A)} The
    two bridge entities, a church and a person, are named only in French
    and only in Serbian, so an agent must enter both languages before the
    branches converge on their shared town and yield the answer.
    \textbf{(B)} The same example under our path-aligned audit: failures in
    building the language cover, in recovering the path and its bridges, and
    after the answer is already exposed.}
    \label{fig:teaser}
\end{figure}

Recent multilingual search-agent benchmarks vary the prompt, task context, or supporting-evidence language~\citep{zhou2025browsecompzh,lee2026kbrowsecomp,almeida2026marca,lu2026xbcp}. While they reveal important language-dependent failures, they do not certify for each item that the agent must recognize an insufficient language environment and gather complementary evidence across languages. Evaluating escape from a language information cocoon therefore requires making the cross-language evidence path part of task construction.

We therefore introduce \method{}, a benchmark that makes the cross-language evidence path an explicit part of task construction. \method{} builds 200 cross-lingual multi-hop tasks from a frozen Wikidata-derived graph and multilingual lexical interface. Different parts of each task's evidence are exposed through different language environments, forming \emph{language-conditioned information islands}. For each task, a construction certificate records the required evidence branches, their bridge entities, the unique answer, and the task's \emph{minimum language cover}: the smallest set of languages needed to provide lexical access to all hidden retrieval-key entities.

Each semantic task is rendered under three parallel versions: English (\texttt{EN}), Chinese (\texttt{ZH}), and a task-specific designated language (\texttt{DL}).
\texttt{EN} serves as the common lingua franca of current language models,
\texttt{ZH} as an unrelated high-resource control language, and \texttt{DL} as
a task-specific language selected from one evidence branch---a language in
which an entity on that branch has a usable surface form.
\texttt{DL} can therefore provide a lexical entry point into one search chain
without necessarily covering the languages needed by the remaining branches.
All three prompts preserve the same answer and underlying construction certificate.
The resulting benchmark is intended to test whether an agent can recognize
that its current language environment is insufficient, cross into another
information island, and integrate evidence across the boundary.

The benchmark remains challenging: mean accuracy across the three prompt modes
ranges from 33.8\% for DeepSeek-V4-Flash to 69.7\% for Claude-Opus-4.8,
leaving even the strongest system below 70\%.

The construction certificate also enables process-level failure-mode analysis.
We introduce a path-aligned failure-mode audit that separates completion of the
minimum language cover from construction of the graph evidence path.
The audit maps an incorrect trajectory to the earliest failed stage: incomplete
or incorrect language routing (L), downstream evidence-chain failure after
minimum-cover completion (T), or residual failure after strong answer evidence
(R).

The analysis yields three main findings.

First, the dominant problem is \emph{multilingual routing rather than language
switching itself}. Agents often leave the prompt language yet enter the
wrong language or stop after visiting only part of the required cover. Most errors occur before a valid minimum language cover is
completed, but simple source-language or English lock-in accounts for only a
small fraction of them. 

Second, designated-language prompts help agents go further, but not always in
the right direction. They produce far fewer incomplete-cover errors and more
trajectories that reach downstream evidence recovery, but they also increase
wrong-language routing. A task-associated language can therefore provide a
useful lexical foothold without identifying the correct multilingual route.

Third, completing the language cover does not complete the search. Even after
reaching the required language environments, agents often fail to recover all
bridge evidence needed across the graph branches. 

Our contributions are:
\begin{itemize}
    \item \textbf{The \method{} benchmark.}
    We construct 200 cross-lingual multi-hop semantic tasks with matched
    \texttt{ZH}/\texttt{EN}/\texttt{DL} prompts. Each item pairs its question
    and answer with certified evidence paths, language requirements, and
    provenance.
    \item \textbf{A verifiable generation framework.}
    We combine a frozen knowledge graph, multilingual lexical coverage,
    branch-necessity constraints, uniqueness checks, and provenance tracking
    to create tasks whose required cross-language evidence routes are
    explicitly certified.
    \item \textbf{A path-aligned diagnosis.}
    The same certificates support process-level analysis that separates
    failures in multilingual routing, evidence-path recovery, and answer use
    after strong evidence exposure.
    \item \textbf{A cross-model empirical study.}
    Evaluation of ten models under three prompt conditions shows that the main
    challenge is not simply crossing a language boundary, but finding and
    completing the right multilingual route before recovering and integrating
    the required evidence.
\end{itemize}

\section{Related Work}
\label{sec:related}

\paragraph{Web search agents and multilingual search benchmarks.}
Benchmarks such as GAIA~\citep{mialon2024gaia} and
BrowseComp~\citep{wei2025browsecomp} evaluate agents that must use external
tools and iteratively search the Web for hard-to-find information.
BrowseComp-Plus~\citep{chen2026browsecompplus} further introduces a fixed,
verified retrieval corpus for more reproducible evaluation of deep-search
agents.
Recent work has extended this setting across languages:
BrowseComp-ZH and K-BrowseComp target Chinese and Korean Web
search~\citep{zhou2025browsecompzh,lee2026kbrowsecomp},
MARCA compares English and Portuguese Web information
seeking~\citep{almeida2026marca},
X-WebAgentBench~\citep{wang2025xwebagentbench} evaluates agents in multilingual
Web environments, and Cross-lingual
BrowseComp-Plus~\citep{lu2026xbcp} varies or distributes supporting evidence
across languages.
These benchmarks expose multilingual search failures, but multilinguality is primarily expressed through the query, interface, or evidence language rather than through a certified cross-language search structure.
\method{} instead constructs each task around an explicit cross-language entity and evidence path, allowing matched prompt interventions and path-aligned diagnosis of search trajectories.

\paragraph{Constructing tasks for search agents.}
Prior work constructs challenging search tasks by controlling the structure and visibility of evidence on the web. WebSailor~\citep{li2025websailor} generates questions through web-grounded
entity--relation graphs, subgraph sampling, and information obfuscation.
WebExplorer~\citep{liu2025webexplorer} combines model-driven web exploration
with iterative removal of salient clues and replacement of explicit references.
WebShaper~\citep{tao2025webshaper} formalizes information-seeking tasks through
compositions, unions, and intersections of Knowledge Projections, using
these structures to guide task expansion. Whereas these methods primarily
control what evidence is exposed and how it is composed, \method{} combines
relational clues that jointly identify a unique answer with complementary
language-specific name coverage, controlling both evidence structure and the
languages through which it can be accessed.

\section{Problem Definition and CocoonBench Generation}
\label{sec:generation}

\begin{figure}[t]
\centering
\includegraphics[width=\linewidth]{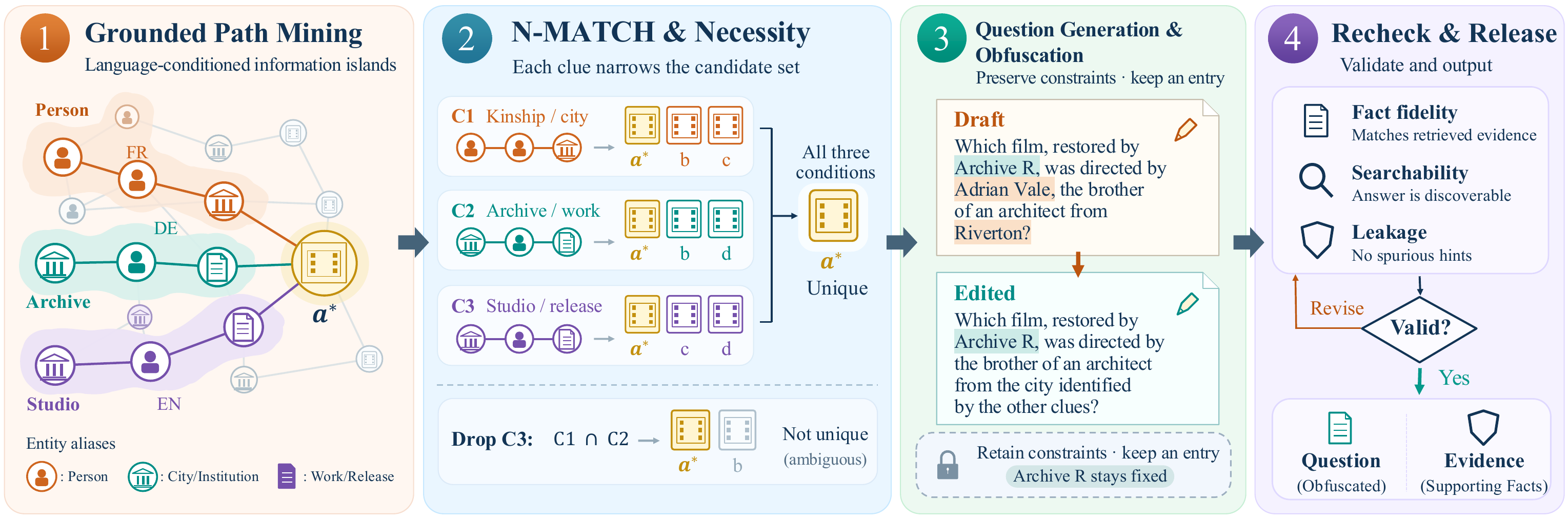}
\caption{
Example generation pipeline for a task asking for a film.
Three relational clues---the director's family tie (C1), the restoring archive (C2), and the studio and release information (C3)---each define a set of candidate films.
Their intersection leaves only the target film $a^*$.
Removing C3 leaves both $a^*$ and $b$ as valid candidates, showing that C3 is necessary.
During verbalization, identifying names are replaced by relational descriptions while selected anchors are retained as search entry points.
}
\label{fig:pipeline}
\end{figure}

\method{} generates cross-language search tasks from a known target entity, ensuring that the clues jointly identify a unique answer (\emph{answer identifiability}), each clue is necessary for that identification (\emph{clue necessity}), and recovering the hidden entities requires complementary lexical access across different languages (\emph{cross-language lexical necessity}).
The resulting structured task is then verbalized into a natural-language question and rendered under matched language conditions.
Finally, we revalidate the task and retain the graph paths, lexical information, provenance, and answer identifiers used during construction as a \emph{construction certificate}.
Figure~\ref{fig:pipeline} illustrates the overall process.

\subsection{Graph Mining with Structural Guarantees}
\label{sec:generation-mining}

We construct each task around a known target $x^*$ and a requested answer $a^*$, which is either the target itself or one of its properties.
From a frozen Wikidata-derived graph~\citep{vrandecic2014wikidata}, we mine $N$ executable relational clues, called \emph{MATCHes}, while preserving relation direction and attached conditions.
One or more explicitly named entities, or \emph{anchors}, are retained as search entry points.
The following subsections formalize the checks used to retain the resulting clue combinations.

\subsubsection{Answer Identifiability}

Each MATCH $i\in\{1,\ldots,N\}$ defines a candidate set $\mathcal{C}_i$ containing all target entities that satisfy its relational constraints.
The selected MATCHes must jointly identify a unique target:

\begin{equation}
\mathcal{C}
=
\bigcap_{i=1}^{N}\mathcal{C}_i
=
\{x^*\}.
\label{eq:generation-intersection}
\end{equation}

Let $F$ map a set of target candidates to the set of answers requested by the question.
When the target itself is the answer, $F$ is the identity mapping; otherwise, $F$ reads the requested property from the target.
In either case, we require $F(\mathcal{C})=\{a^*\}$.

\subsubsection{Clue Necessity}

A unique answer does not imply that every clue is useful.
A task could appear multi-clue even if one or more clues can be removed without changing the answer.
We therefore require every retained MATCH to be individually necessary.

For each MATCH $i$, we remove it and rerun the remaining executable constraints.
The answer must become ambiguous:

\begin{equation}
\left|
F\left(
\bigcap_{j\neq i}\mathcal{C}_j
\right)
\right|
> 1
\qquad
\text{for every } i.
\label{eq:generation-match-necessity}
\end{equation}

Here, $j$ indexes the remaining MATCHes and $|\cdot|$ denotes set cardinality.
We apply the same deletion test to individual factual constraints within a MATCH.
After each deletion, the complete query is rerun while preserving the remaining relations, shared variables, and answer relation. For example, in the \v{C}elinac task in Figure~\ref{fig:teaser}, the church and Iguman Antonije branches form two separate MATCHes whose candidate sets intersect only at \v{C}elinac.

\subsubsection{Cross-Language Lexical Necessity}

The Wikidata graph itself is language-neutral. However, some entities only have names in specific languages. Thus, the cross-language requirement comes from the lexical interfaces through which hidden entities can be discovered.

We therefore ask whether a single language can provide usable lexical access to all hidden retrieval entities required by the task.
Let $R$ denote this set of hidden retrieval-key entities, excluding named anchors and the final answer.
For an entity $v\in R$, let $\Lambda(v)$ denote the set of supported languages in which the frozen lexical inventory provides a usable label or alias.
Let $\mathcal{L}$ be the set of supported languages.
A candidate set $L\subseteq\mathcal{L}$ covers the task if every entity in $R$ is named in at least one language in $L$, and $|L|$ is the number of languages it contains.
We define the minimum lexical-cover size as

\begin{equation}
\tau_{\mathrm{lex}}(R)
=
\min_{L\subseteq\mathcal{L}}
\left\{
|L| :
\forall v\in R,\;
L\cap\Lambda(v)\neq\varnothing
\right\},
\label{eq:generation-language-cover}
\end{equation}

We retain tasks with
$\tau_{\mathrm{lex}}(R)\geq2$:
at least two languages are therefore needed to provide lexical access to all required hidden entities under the frozen inventory.
This minimum lexical-cover size is the task's \emph{minimum language cover}.
To prevent this requirement from being satisfied trivially through the fixed Chinese or English prompt conditions, entities counted toward the multilingual requirement must lack usable Chinese or English labels or aliases in the frozen inventory.

In the \v{C}elinac task, the certificate-designated bridge entities are Saint-Gabriel Church, with a usable French surface, and Iguman Antonije, with a usable Serbian surface.
Neither language names both entities, so the minimum cover is French plus Serbian.

\subsection{Question Realization}
\label{sec:generation-rendering}

\paragraph{Verbalization and obfuscation.}

After the graph structure is fixed, we convert it into a natural-language question while preserving the underlying facts, relation directions, answer relation, and entity roles. During verbalization, hidden entities are replaced with relational descriptions while selected anchors remain explicitly named as search entry points.
We remove direct identifiers and overly identifying details that would make the answer recoverable through a trivial lookup, while preserving the relational constraints required by the executable task.
This follows the general use of identifier obfuscation in search-task generation~\citep{li2025websailor,liu2025webexplorer}.
Any edit that changes a factual constraint triggers a fresh execution of the uniqueness and necessity checks.

\paragraph{Matched prompt conditions.}

Each semantic task is rendered under three matched prompt conditions:
Chinese (\texttt{ZH}), English (\texttt{EN}), and a task-specific designated language (\texttt{DL}).
\texttt{ZH} provides an unrelated high-resource prompt condition, \texttt{EN} provides a common high-resource baseline, and \texttt{DL} is a language selected from one evidence branch---a language in which an entity on that branch has a usable surface form. The three prompts preserve the same semantic and are parallel.
The \texttt{DL} condition may therefore provide a useful lexical entry point into one search chain without necessarily covering the remaining branches.

\subsection{Validation and Construction Certificates}
\label{sec:generation-validation}

\paragraph{Task validation.}

We validate each generated task at both the structural and factual levels.
Structural validation reruns the unique-answer and clue-necessity checks, including constraint-level deletion tests, after any modification to the task. For factual validation, we start from the reference answer and trace every claim in the final question back to its supporting evidence.
We additionally check that the retained clues remain searchable without exposing the answer, Wikidata identifiers, or other direct shortcuts.
Quality control combines these checks with the fixed-query shortcut audit and
human verbalization review detailed in Appendix~\ref{app:quality}.
For each candidate, the shortcut audit issues translated full-question and
branch queries through the same Serper backend used in evaluation, using
translated relation terms and transliterated public anchors.  A task is
removed whenever an audited single-language query exposes the correct answer
or an equivalent alias.  Every retained task therefore passes the predefined
single-language shortcut screen.
Tasks that fail any check are revised and revalidated.

\paragraph{Construction certificates.}

For every retained task, we store the information produced during generation as a \emph{construction certificate}.
The certificate includes the executable MATCHes, graph paths and shared variables, lexical inventories and language covers, source provenance, bridge entities, target and answer identifiers, and exact answer Wikipedia URLs. A \emph{bridge entity} is a certificate-designated entry into one evidence branch.
These records provide the observable signals used by the trajectory audit in Section~\ref{sec:audit}.

\paragraph{Benchmark composition.}

\method{} contains 200 semantic tasks.
Rendering each task under the three prompt conditions yields 600 cases per evaluated system, and the \texttt{DL} condition spans 33 languages; Appendix~\ref{app:dl-coverage} reports the full language, language-pair, writing-system, cover-size, and per-language result distributions. Of the 200 tasks, 170 are constructed from two executable relational clues (MATCHes; 85\%), 28 from three (14\%), and two from four (1\%).
The maximum clue-path depth within a task ranges from 2 to 8, with a median of 3.
Summing clue-path depths within each task and averaging across the 200 tasks gives an average total path depth of 6.36.

\section{Evaluation}
\label{sec:evaluation}

\label{sec:setup}

All models use the same agent harness and Google Search through the Serper API.
Tool access, webpage parsing and summarization, prompts, grading, inference budgets, and exception handling are detailed in Appendix~\ref{app:agent-setup}.

We evaluate CocoonBench on 10 state-of-the-art models: Claude-Opus-4.8 (Max)~\citep{anthropic2026claudeopus48}, GPT-5.6-Sol (Max)~\citep{openai2026gpt56}, GPT-5.5 (Extra High)~\citep{openai2026gpt55}, Kimi-K3 (Max)~\citep{kimiteam2026kimik3}, Qwen3.8-Max (Extra High)~\citep{qwenteam2026qwen38max}, GLM-5.3-Flash (Max)~\citep{zai2026glm53flash}, GLM-5.3 (Max)~\citep{zai2026glm53}, GLM-5.2 (Max)~\citep{zai2026glm52}, DeepSeek-V4-Pro-0813 (Max), and DeepSeek-V4-Flash-0731 (Max)~\citep{deepseekai2026deepseekv4}. All models are accessed through their official API providers, and all requests and responses are handled in compliance with the corresponding terms of use.

\begin{figure}[t]
    \centering
    \includegraphics[width=\linewidth]{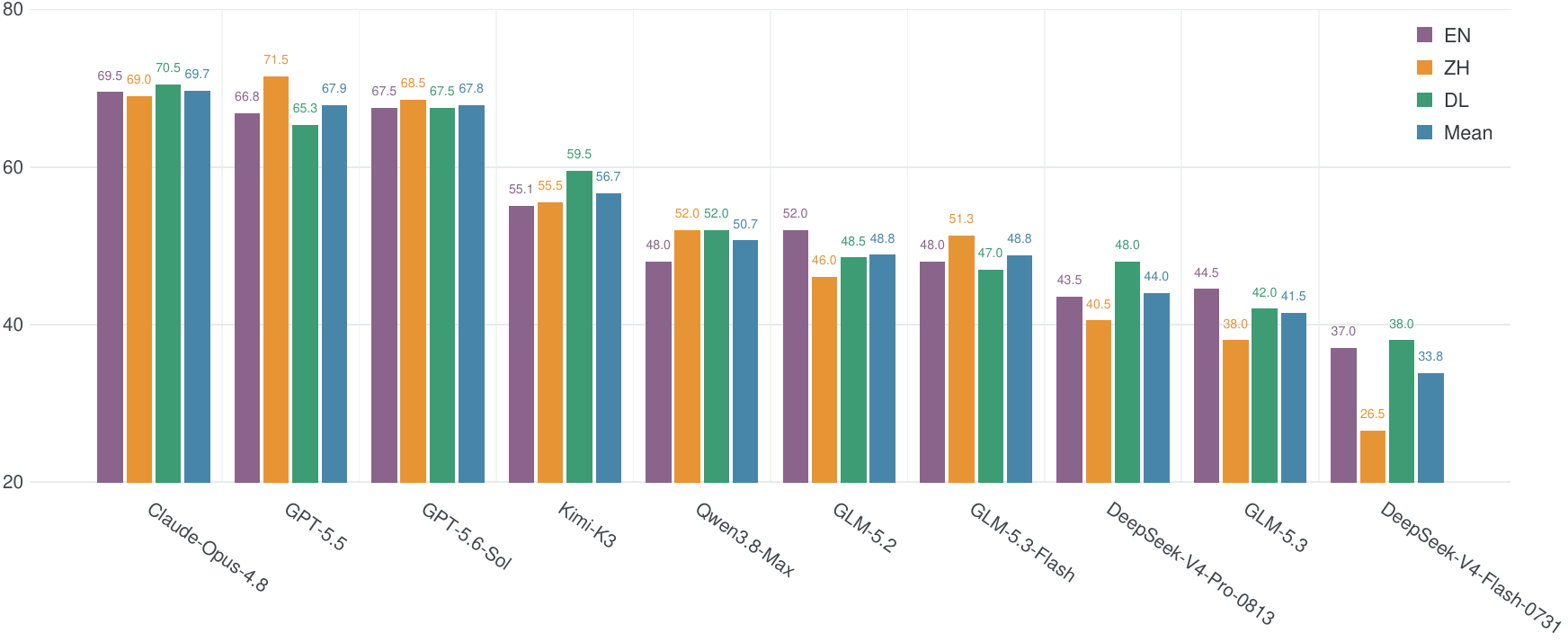}
    \caption{Performance of state-of-the-art LLMs on \method{} under three prompt modes:
    \texttt{EN} (purple), \texttt{ZH} (orange), and \texttt{DL} (green). The blue columns represent the mean score of the three modes. Models are ranked by their mean scores.}
    \label{fig:performance}
\end{figure}

\label{sec:benchmark-results}
\label{sec:performance}
Figure~\ref{fig:performance} shows the performance of state-of-the-art LLMs on CocoonBench under three prompting language modes: \texttt{ZH}, \texttt{EN}, and \texttt{DL}. The agent user prompt (see Figure~\ref{fig:prompt-agent-user}) comprises an instruction and a question; the three modes use the corresponding fixed Chinese, English, and designated-language renderings produced during benchmark construction.

Across the three prompt modes, model mean accuracy spans 35.9 percentage points, from 33.8\% (DeepSeek-V4-Flash) to 69.7\% (Claude-Opus-4.8). This spread shows that CocoonBench separates the multilingual search-agent capabilities of current leading models with no clear sign of saturation. At the same time, the three prompt modes yield substantial performance differences on most evaluated models, underscoring the language gap in cross-lingual search-agent tasks.

Examining the three modes more closely, neither English nor Chinese emerges as the dominant language across all tested models, suggesting that there is no universally ``best'' language for solving CocoonBench problems. In fact, the gap between the best and worst prompt modes reaches 11.5\,pp (DeepSeek-V4-Flash), indicating severe performance degradation under sub-optimal language choices.

\section{Failure-Mode Analysis}
\label{sec:failure-mode-analysis}

We use each task's construction certificate to classify incorrect search
trajectories; the following subsections define the path-aligned audit and
report its findings.

\subsection{Path-Aligned Failure-Mode Audit}
\label{sec:audit}

Each \method{} item's construction certificate records a set of languages that
must be visited, bridge entities for the required graph branches, entities
along the evidence paths, and the final answer entity. We use these annotations to locate where an
incorrect search trajectory breaks down. In particular, we distinguish failures
to reach the required language environments (L), failures to recover the
evidence path after doing so (T), and failures that occur even after strong
evidence for the answer has already appeared (R).

\paragraph{Observable signals.}

We identify the language of each WebSearch query using
fastText~\citep{joulin2017bag,joulin2017fasttext}. For Wikipedia search results
and fetched pages, we infer the language from the Wikipedia hostname. We count a
language as \emph{entered} only when the agent issues a query in that language
or fetches a page in that language. For trajectory auditing, a minimum language
cover is complete once all of its required languages have been entered. We
detect path and bridge entities in the trajectory in three ways: through an
explicit QID, a Wikipedia URL that resolves to that QID, or a normalized
multilingual surface form of the entity.

\begin{table}[t]
  \caption{Operational definitions of the three failure stages: Language
  Covering (L), Evidence-Path Recovery (T), and Residual Failure after Answer
  Exposure (R).}
  \label{tab:ltr}
  \centering
  \scriptsize
  \setlength{\tabcolsep}{2.2pt}
  \begin{tabular}{@{}p{0.1\linewidth}p{0.2\linewidth}p{0.325\linewidth}p{0.325\linewidth}@{}}
  \toprule
  \textbf{Stage} & \textbf{Subtype} & \textbf{Meaning} &
  \textbf{Decision rule} \\
  \midrule

  Language Covering (L)
  & L1 source lock
  & A non-English run never leaves the prompt language.
  & For a non-English prompt, no identifiable query or page fetch is made in
  any other language. \\

  \cmidrule(lr){2-4}

  & L2 English trap
  & An English run remains trapped around English, or a non-English run falls
  back to English without non-source valid-route progress.
  & English is entered, no required non-source language on a valid route is
  entered, and no sustained non-route-language branch is followed. \\

  \cmidrule(lr){2-4}

  & L3 wrong route
  & The agent switches languages, but to a language that is not useful for any
  valid route.
  & The run leaves the prompt language, but reaches no required non-source
  language that belongs to a valid cover. \\

  \cmidrule(lr){2-4}

  & L4 incomplete cover
  & The agent enters at least one required cover language beyond the prompt
  language, but does not complete the cover.
  & At least one required non-source cover language is entered, but the runtime
  cover is not completed. \\

  \midrule

  Evidence-Path Recovery (T)
  & T1 path-entry failure
  & The required language cover is completed, but the retrieved content does
  not reach the annotated evidence path.
  & The required language cover is completed, but no configured path entity is detected in
  the retrieved results or pages. \\

  \cmidrule(lr){2-4}

  & T2 incomplete bridge recovery
  & The required language cover is completed, but one or more evidence branches
  remain uncovered by their designated bridge entities.
  & At least one path entity is detected, but fewer than all required branches
  are covered by a designated bridge entity. \\

  \cmidrule(lr){2-4}

  & T3 answer-evidence failure
  & Every required evidence branch is covered by a designated bridge entity,
  but strong evidence for the final answer never appears.
  & All required branches are covered, but neither the answer QID nor the exact
  answer Wikipedia URL appears. \\

  \midrule

  Residual Failure after Answer Exposure (R)
  & R1 fetched-page exposure
  & Strong answer evidence appears in a page that the agent opens, but the run ends without a correct completion.
  & All previous gates are passed, and a fetched page exposes the answer QID or
  exact answer Wikipedia URL. \\

  \cmidrule(lr){2-4}

  & R2 search-result-only exposure
  & Strong answer evidence appears in the search results, but the answer page is
  never opened.
  & All previous gates are passed, the answer QID or exact answer Wikipedia URL
  appears in search results, and no fetched page contains strong answer
  evidence. \\

  \bottomrule
  \end{tabular}
\end{table}

\subsection{Analysis}

\subsubsection{Most Failures Come from Incomplete Multilingual Routing}

\label{sec:dominant}

Language-covering failures dominate the classifiable errors. They account for
65.7\% (L), more than twice the 32.3\% that occur during evidence-path recovery
after cover completion (T); 1.9\% occur after strong answer evidence has
already appeared (R).

Importantly, these failures are rarely caused by the agent simply staying in
one language. Source lock and English trap (L1/L2) together account for only
5.8\% of errors. Much more often, the agent does switch languages but fails to
follow the right multilingual route. In 22.2\% of errors, it leaves the prompt
language but moves into the wrong language (L3). Another 37.8\% reach some of
the required cover languages beyond the prompt language but stop before
completing the cover (L4), making
incomplete cover the largest individual failure type in
Figure~\ref{fig:failure}.

The main difficulty is therefore not getting agents to switch languages at
all, but getting them to visit the full set of languages needed to recover the
relevant evidence.

\subsubsection{Designated-Language Prompts Help Agents Go Further, but Not Always in the Right Direction}

\label{sec:pseudobridge}

The designated-language (\texttt{DL}) prompt gives a small average accuracy
gain in Figure~\ref{fig:performance}, suggesting that starting from a language
selected from one evidence branch can provide a useful entry point for
search.

The error patterns show a similar effect. Among classifiable errors,
\texttt{DL} produces far fewer incomplete-cover cases than \texttt{ZH} or
\texttt{EN} (13.5\% versus 51.3\%/51.4\%), and more of its errors reach the
later T stage (43.6\% versus 22.0\%/29.1\%). In this sense, the designated
language often helps the agent make more progress through the search.

However, that progress is not always in the right direction. Wrong-language
routes are much more common under \texttt{DL}
(34.3\% versus 16.7\%/14.4\%). A language selected from one evidence branch can therefore
serve as a useful foothold while also anchoring the search to a locally
plausible but ultimately incorrect route.

\subsubsection{Completing the Language Cover Does Not Complete the Search}

\label{sec:residual}

Reaching all required languages is only the first part of the problem. Even
after a valid language cover is complete, agents can still fail to recover the
evidence needed to reach the answer.

Among all classifiable errors, 9.6\% complete the language cover but never
reach the configured evidence path (T1). Another 19.3\% reach the path but
cover none or only some of the required evidence branches through designated
bridge entities (T2), making incomplete bridge recovery the largest post-cover
failure mode. A further 3.5\% cover every required evidence branch through a
designated bridge entity but still fail to expose strong evidence for the
answer (T3).

Only a small number of errors remain after strong answer evidence itself has
already appeared. These residual R failures account for 1.9\% of all
classifiable errors. They are split almost evenly between cases where the
answer evidence appears in a fetched page (47.8\%, R1) and cases where it
appears only in the search results (52.2\%, R2).

Thus, simply reaching the right languages is not enough: the agent must still
follow the evidence path, recover the required bridges, and make use of the
answer evidence it encounters. Appendix~\ref{app:subtype-cases} gives examples
of both residual exposure patterns.

\begin{figure}[t]
    \centering
    \includegraphics[width=\linewidth]{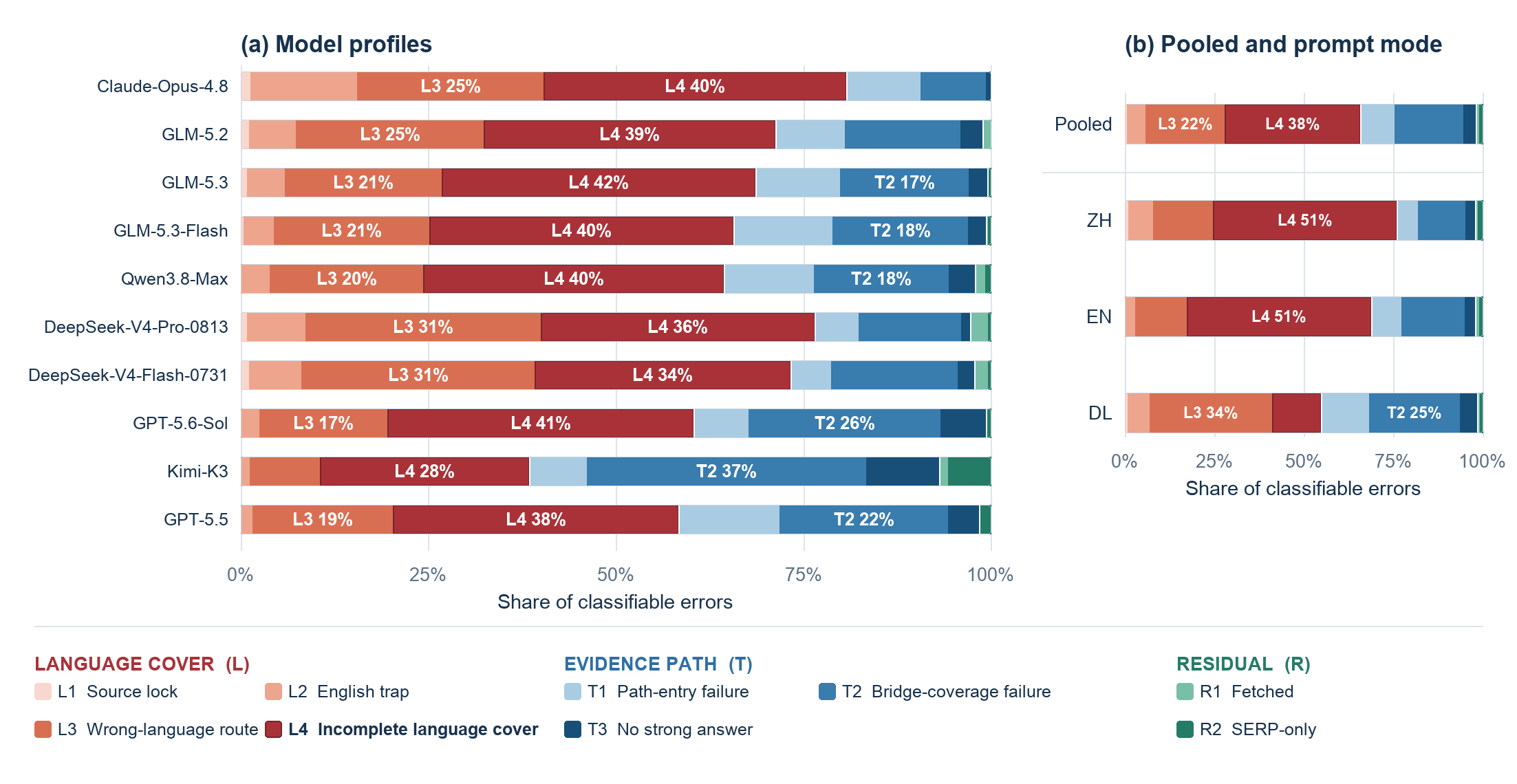}
    \caption{Failure-category percentage composition of incorrect, classifiable
    trajectories. Colors group failures into
    language-cover routing (L), evidence-path recovery (T), and residual
    failure (R); the left panel groups errors by model and the right panel by
    prompt mode.}
    \label{fig:failure}
\end{figure}

\subsubsection{Different Models Break Down at Different Stages}
\label{sec:profiles}

The overall error distribution hides substantial differences across models.
Claude-Opus-4.8 and the two DeepSeek variants fail predominantly before completing a valid language cover, with language-covering failures (L) accounting for 80.7\%, 73.3\%, and 76.5\% of their classifiable errors, respectively.
Kimi-K3 shows the opposite pattern: evidence-path recovery failures (T) form its largest error family at 54.7\%, while GPT-5.5 and GPT-5.6-Sol also fail relatively late, with 40.1\% and 39.0\% of their errors reaching the evidence-path recovery stage.
These systems more often reach the required languages but still fail to recover a complete chain of path and bridge evidence.
Qwen3.8-Max and the three GLM variants fall between these two profiles: their errors are still dominated by language-covering failures, but 27.6\%--33.7\% reach the evidence-path recovery stage.
Residual failures after answer exposure (R) are rare overall, although Kimi-K3 has the largest share at 6.9\%.
Together, these results show that models do not simply differ in how often they fail, but also in how far they typically progress before failing.

\section{Conclusion}
\label{sec:conclusion}

\method{} certifies cross-language evidence requirements during task
construction and reuses the resulting construction certificates for both
final-answer evaluation and path-aligned trajectory diagnosis.

Across ten model--agent systems, multilingual capability did not ensure
effective multilingual search: agents often failed either to complete the
required language route or to turn language coverage into a verifiable
evidence chain.

The answer to our title question is therefore: only partially. Escaping a
language information cocoon requires agents to recognize missing coverage,
choose the next language strategically, and sustain branch-aware retrieval
until the evidence chain is complete.

\section*{AI Use Statement}

Generative AI tools were used to polish the writing of this paper, and also to 
generate synthetic datasets; the corresponding procedures are
described in the paper. The authors take full responsibility for the final content of this
work.

\newpage
\bibliography{references}
\bibliographystyle{iclr2025_conference}

\newpage
\appendix

\section{Discussion}
\label{sec:discussion}

\paragraph{Language Shapes the Agent's Visible World.}

For a search agent, language is not merely the surface form of a query.
It also shapes which entities, pages, and evidence are likely to become visible during search.
The same Web can therefore present different effective information spaces depending on the language through which it is accessed.
This perspective helps explain the language information cocoon: an agent may have access to the global Web while effectively operating within only the portion exposed by its current search language.
Multilingual search requires not only understanding information across languages, but also recognizing when the currently visible world is incomplete.

\paragraph{Multilinguality in the Model Does Not Guarantee Multilinguality in Search.}

Modern language models can understand and generate many languages~\citep{zhu2024multilingual,zeng2025linguafranca,chan2026xtom}, yet this capability
does not automatically translate into multilingual search behavior.
Our results are consistent with a gap between multilingual language competence and multilingual search policy.
Agents may be able to read evidence in another language once they encounter it, while still failing to complete the minimum language cover, track which required languages have actually been queried or fetched, or maintain branch-aware search after route completion.
Among classifiable errors, \texttt{ZH} and \texttt{EN} have almost identical incomplete-cover shares (51.3\% and 51.4\%), so high-resource English is not exempt from the route-completion bottleneck; \texttt{EN} errors more often reach downstream path stages, but these conditional compositions should not be interpreted as causal effects of prompt language.
A model can therefore be multilingual at the level of representation and generation while still using languages ineffectively during search~\citep{han2025headsup}.
Improving multilingual agents may require explicit route and coverage tracking together with branch-aware search control, rather than assuming that multilingual capability will transfer automatically to agentic search.

\paragraph{Language Barriers May Amplify Information Inequality.}

These findings also have implications for access to information.
If search agents remain biased toward the language of the user's query, users may receive different subsets of the available Web depending on the language they speak.
This matters especially for lower-resource languages, where less information may be available locally and useful evidence may more often need to be retrieved from another language environment~\citep{bao2012omnipedia,wulczyn2016growing,roy2020information,zeng2024mbs}.
Users whose languages are already underrepresented on the Web could therefore be disproportionately affected by weak cross-language search.
More broadly, multilingual search quality is not only a question of whether an agent can answer in a user's language, but also whether users across languages are given comparable access to the evidence available on the wider Web.

\section{Agent Setup}
\label{app:agent-setup}

\paragraph{Tools and Web access.}
All models use the same agent harness with four tools: \texttt{WebSearch}, \texttt{WebFetch}, \texttt{Bash}, and \texttt{ExecuteCode}. The first two let the agent search for and read Web pages, while the latter two provide a terminal session and a Python environment with a Jupyter kernel. \texttt{WebSearch} uses Google through the Serper API and returns 10 results per query. For \texttt{WebFetch}, Jina parses retrieved HTML into Markdown with up to 200{,}000 characters in the main body. DeepSeek-V4-Flash-0731~\citep{deepseekai2026deepseekv4} then summarizes the parsed content using the prompt in Figure~\ref{fig:prompt-summary}; tool-result truncation controls the summary length.

\paragraph{Agent prompt format.}
The system and user prompt templates are shown in Figures~\ref{fig:prompt-agent-system} and~\ref{fig:prompt-agent-user} in Appendix~\ref{app:prompts}. The \texttt{\{today\_date\}} placeholder in the system prompt is replaced by the evaluation date.
For each semantic task, the \texttt{ZH}, \texttt{EN}, and \texttt{DL} modes
use the corresponding fixed instruction and question renderings produced
during benchmark construction.

\paragraph{Answer parsing and grading.}
The agent submits its final answer string through a \texttt{finish} call, and a judge model, which is Gemini-3.5-Flash \citep{kavukcuoglu2026gemini35} in our setup, is prompted to decide whether the agent's answer is equivalent to the ground truth. If the agent never issues a \texttt{finish} call, the last content in its trajectory is used as the final answer. The judging prompts are shown in Figures~\ref{fig:prompt-judge-system} and~\ref{fig:prompt-judge-user}. The judgment is extracted with the regular expression \texttt{<is\_equivalent>\textbackslash s*(.*?)\textbackslash s*</is\_equivalent>}, returning $1.0$ for equivalent answers and $0.0$ otherwise.

\paragraph{Inference budgets.}
For each question, we allow at most 1000 turns of response and tool calling and a 6-hour execution limit, with a 60-second timeout per tool call. To tolerate occasional errors, we set the maximum number of retries to 5 for each of the main client network/service, LLM response, tool arguments, and tool calling. The maximum context size is 65{,}536 tokens; reaching it triggers compression of intermediate search and tool-call results. Tool-call results are truncated to 10{,}000 characters before being fed back into the response, while the maximum output length is 128K tokens under the highest reasoning effort. Hyper-parameters such as top-$k$, top-$p$, and temperature are kept at their defaults for all evaluated models.

\paragraph{Exception handling.}
When the agent, judge model, or tools raise timeout or network/service errors that are not resolved by automatic retries, the sample is re-run up to 20 times, and the first successful result is accepted. Any remaining timeout and refusal-to-answer errors are then counted as incorrect answers, while all other remaining errors are skipped in scoring.

\section{Benchmark Quality and Human Evaluation}
\label{app:quality}

This appendix evaluates the quality of \method{} beyond model performance.
We report the maintenance of the final Gold set, human judgments of
verbalization fidelity, language identification, and the accompanying release
artifacts.

\subsection{Gold-Set Repair and Translation Validation}

The Gold-set repair starts from 200 tasks.  We remove 11 tasks with confirmed
quality issues and retain 189, then add 11 replacements from the historical
candidate pool to form the final 200 tasks.  All 200 tasks are subsequently
translated into \texttt{EN} and \texttt{DL}.  Translation batches are retried
until every task passes the automatic structural contract checks.  The final
construction certificates record answer-identifiability and clue-necessity
proofs for all 200 tasks.

\begin{table}[t]
\caption{Maintenance history of the final 200-task Gold set.}
\label{tab:quality-filtering}
\centering
\small
\setlength{\tabcolsep}{5pt}
\begin{tabular}{@{}lrrr@{}}
\toprule
\textbf{Stage} & \textbf{Removed} & \textbf{Added} &
\textbf{Remaining} \\
\midrule
Initial Gold set & -- & -- & 200 \\
Confirmed problematic tasks removed & 11 & -- & 189 \\
Historical replacement tasks added & -- & 11 & 200 \\
Automatic translation and contract validation & 0 & -- & 200 \\
\bottomrule
\end{tabular}
\end{table}

\subsection{Designated-Language Coverage}
\label{app:dl-coverage}

The 200 \texttt{DL} renderings span 33 designated languages.
Table~\ref{tab:dl-language-results} reports each language's share of semantic
tasks, pooled \texttt{DL}-mode accuracy across the ten systems, and L/T/R
composition among its incorrect, classifiable trajectories.  Task share uses
the 200 semantic tasks as denominator; accuracy uses all \texttt{DL}
model--task runs assigned to that language.

\begin{table}[t]
\caption{Designated-language distribution and per-language results.
``Share'' is the percentage of semantic tasks assigned to each language;
accuracy is pooled across the ten evaluated systems.  L/T/R percentages use
the classifiable errors for that language.}
\label{tab:dl-language-results}
\centering
\scriptsize
\setlength{\tabcolsep}{4.0pt}
\begin{tabular}{@{}lrrrrr@{}}
\toprule
\textbf{Designated language} & \textbf{Share} & \textbf{Accuracy} &
\textbf{L} & \textbf{T} & \textbf{R} \\
\midrule
Lithuanian (\texttt{lt}) & 11.0 & 45.5 & 38.1 & 59.3 & 2.7 \\
Dutch (\texttt{nl}) & 6.5 & 48.5 & 58.7 & 39.7 & 1.6 \\
Russian (\texttt{ru}) & 6.0 & 56.7 & 81.6 & 16.3 & 2.0 \\
German (\texttt{de}) & 6.0 & 41.7 & 32.3 & 63.1 & 4.6 \\
Czech (\texttt{cs}) & 6.0 & 43.3 & 63.5 & 34.9 & 1.6 \\
Hebrew (\texttt{he}) & 5.0 & 51.0 & 48.9 & 48.9 & 2.1 \\
Arabic (\texttt{ar}) & 4.5 & 46.7 & 63.4 & 36.6 & 0.0 \\
Ukrainian (\texttt{uk}) & 4.5 & 53.3 & 77.5 & 22.5 & 0.0 \\
Persian (\texttt{fa}) & 4.0 & 70.0 & 78.9 & 21.1 & 0.0 \\
Turkish (\texttt{tr}) & 4.0 & 82.5 & 69.2 & 30.8 & 0.0 \\
Spanish (\texttt{es}) & 4.0 & 62.5 & 57.7 & 42.3 & 0.0 \\
French (\texttt{fr}) & 4.0 & 66.2 & 60.0 & 40.0 & 0.0 \\
Italian (\texttt{it}) & 3.5 & 70.0 & 61.9 & 38.1 & 0.0 \\
Polish (\texttt{pl}) & 3.5 & 34.3 & 55.6 & 44.4 & 0.0 \\
Slovenian (\texttt{sl}) & 3.0 & 56.7 & 58.3 & 41.7 & 0.0 \\
Finnish (\texttt{fi}) & 3.0 & 63.3 & 95.5 & 4.5 & 0.0 \\
Hungarian (\texttt{hu}) & 3.0 & 70.0 & 38.9 & 61.1 & 0.0 \\
Catalan (\texttt{ca}) & 3.0 & 26.7 & 32.4 & 64.9 & 2.7 \\
Swedish (\texttt{sv}) & 2.5 & 88.0 & 50.0 & 50.0 & 0.0 \\
Serbian (\texttt{sr}) & 1.5 & 33.3 & 72.2 & 22.2 & 5.6 \\
Romanian (\texttt{ro}) & 1.5 & 83.3 & 0.0 & 75.0 & 25.0 \\
Korean (\texttt{ko}) & 1.0 & 20.0 & 40.0 & 60.0 & 0.0 \\
Bulgarian (\texttt{bg}) & 1.0 & 35.0 & 30.0 & 70.0 & 0.0 \\
Danish (\texttt{da}) & 1.0 & 20.0 & 31.2 & 68.8 & 0.0 \\
Latvian (\texttt{lv}) & 1.0 & 55.0 & 25.0 & 62.5 & 12.5 \\
Greek (\texttt{el}) & 1.0 & 30.0 & 100.0 & 0.0 & 0.0 \\
Slovak (\texttt{sk}) & 1.0 & 75.0 & 0.0 & 100.0 & 0.0 \\
Portuguese (\texttt{pt}) & 1.0 & 20.0 & 37.5 & 62.5 & 0.0 \\
Japanese (\texttt{ja}) & 1.0 & 85.0 & -- & -- & -- \\
Norwegian Bokm{\aa}l (\texttt{nb}) & 0.5 & 80.0 & 0.0 & 100.0 & 0.0 \\
Estonian (\texttt{et}) & 0.5 & 60.0 & 100.0 & 0.0 & 0.0 \\
Indonesian (\texttt{id}) & 0.5 & 30.0 & 71.4 & 28.6 & 0.0 \\
Croatian (\texttt{hr}) & 0.5 & 80.0 & 100.0 & 0.0 & 0.0 \\
\bottomrule
\end{tabular}
\end{table}

\paragraph{Language pairs.}
Two-language minimum covers account for 80.5\% of tasks and span 94 distinct
language pairs.  Table~\ref{tab:dl-language-pairs} reports the most frequent
pairs; all remaining two-language configurations jointly account for 54.5\%
of tasks.

\begin{table}[t]
\caption{Most frequent canonical minimum-language covers.}
\label{tab:dl-language-pairs}
\centering
\small
\setlength{\tabcolsep}{5.0pt}
\begin{tabular}{@{}lr@{}}
\toprule
\textbf{Language pair} & \textbf{Task share (\%)} \\
\midrule
\texttt{lt}--\texttt{ru} & 8.0 \\
\texttt{nl}--\texttt{uk} & 6.0 \\
\texttt{pl}--\texttt{uk} & 3.0 \\
\texttt{ar}--\texttt{fa} & 2.5 \\
\texttt{ca}--\texttt{es} & 2.0 \\
\texttt{he}--\texttt{ru} & 1.5 \\
\texttt{ar}--\texttt{he} & 1.5 \\
\texttt{cs}--\texttt{de} & 1.5 \\
Other two-language pairs & 54.5 \\
Cover size greater than two & 19.5 \\
\bottomrule
\end{tabular}
\end{table}

\paragraph{Writing systems and cover size.}
Table~\ref{tab:dl-structural-distribution} groups designated languages by
their primary writing system and minimum covers by their number of languages.

\begin{table}[t]
\caption{Structural distribution of the 200 semantic tasks.}
\label{tab:dl-structural-distribution}
\centering
\scriptsize
\setlength{\tabcolsep}{4.0pt}
\begin{tabular}{@{}llr@{}}
\toprule
\textbf{Dimension} & \textbf{Category} & \textbf{Task share (\%)} \\
\midrule
\multirow{7}{*}{Writing system}
  & Latin & 70.5 \\
  & Cyrillic & 13.0 \\
  & Arabic & 8.5 \\
  & Hebrew & 5.0 \\
  & Hangul & 1.0 \\
  & Greek & 1.0 \\
  & Japanese & 1.0 \\
\midrule
\multirow{6}{*}{Minimum-cover size}
  & 2 languages & 80.5 \\
  & 3 languages & 7.5 \\
  & 4 languages & 3.5 \\
  & 5--6 languages & 5.0 \\
  & 7--8 languages & 2.5 \\
  & 9 or more languages & 1.0 \\
\bottomrule
\end{tabular}
\end{table}

\subsection{Search-Result Shortcut Audit}
\label{app:serp-shortcut-protocol}

\paragraph{Search conditions.}
We use the same Google Search backend through the Serper API as in model
evaluation and retain the results for every query.  For a
task with $N$ evidence branches, we audit Chinese, English, and every language
in its minimum cover.  Each language receives a set of $N+1$ queries:
one translation of the complete question and one query containing the
description of each individual branch.

All queries are generated from the structured task representation.  They may translate relation terms and transliterate public
anchors already present in the question.  Each single-language query draws
from this predefined set of relation terms, public anchors, and structured
branch descriptions.

\paragraph{Decision rule.}
The audit operates on the titles, snippets, and URLs of the search results.
We normalize these three fields and match them against the reference answer
and all answer aliases stored in the construction certificate.  An answer or
unambiguous equivalent-alias match assigns the task a shortcut finding.  A
zero-match record across the complete query set assigns the task a pass.

\paragraph{Cross-language confirmation.}
We then run one branch query in each language of the minimum cover.  A term
exposed in one language's result page may be used as the handoff for the next
language.  Each handoff uses terms exposed by the preceding result page.  The
route is confirmed when the answer first appears in a title, snippet, or URL
after the language transition.  Appendix~\ref{app:web-scope-stress} shows the
complete process: Latvian results identify Vyatlag, and a Russian follow-up
exposes the modern administrative-center name \emph{Lesnoy}.

\subsection{Human Verbalization Evaluation}

We check the Chinese (\texttt{ZH}) verbalization of every one of the 200
semantic tasks.  Annotators compare each Chinese rendering directly with its
structured task representation, reference answer, and construction
certificate.  Each rendering receives one review, and flagged or uncertain
cases receive a second review.  After a Chinese verbalization is accepted, it
is translated into English (\texttt{EN}) and the task-specific designated
language (\texttt{DL}).  The three renderings preserve a shared answer and
evidence structure across languages.

We evaluate each Chinese verbalization for:
\begin{enumerate}
    \item preservation of all entities, relations, qualifiers, and coreference;
    \item preservation of the reference answer and requested answer type;
    \item complete and exact preservation of all clues;
    \item preservation of which entities are named and which remain hidden;
    \item preservation of the intended answer-cue boundaries; and
    \item grammaticality and fluency.
\end{enumerate}

Table~\ref{tab:translation-quality} reports verbalization fidelity on the
complete set of 200 Chinese renderings.

\begin{table}[t]
\caption{Human evaluation of Chinese verbalization fidelity.  Fluency is
rated from 1 to 5; the remaining columns report pass rates in percent.}
\label{tab:translation-quality}
\centering
\scriptsize
\setlength{\tabcolsep}{3.2pt}
\resizebox{\textwidth}{!}{%
\begin{tabular}{@{}lrrrrr@{}}
\toprule
\textbf{$n$} & \textbf{Semantic equivalence} &
\textbf{Constraint preservation} & \textbf{Answer preservation} &
\textbf{Answer-cue preservation} & \textbf{Fluency} \\
\midrule
200 & 99.0 & 98.0 &
100.0 & 100.0 & 4.72 \\
\bottomrule
\end{tabular}
}
\end{table}

Chinese renderings requiring revision undergo another review cycle before the
accepted text is translated into \texttt{EN} and \texttt{DL}.

We additionally sample 10\% of the \texttt{EN} renderings and 10\% of the
\texttt{DL} renderings.  Human experts proficient in the corresponding
language evaluate them against the accepted Chinese rendering and the same
semantic-fidelity criteria above.  The initial pass rates are 95.0\% for
\texttt{EN} and 90.0\% for \texttt{DL}, with an overall pass rate of 92.5\%.
All flagged translations are corrected and pass a second review.

\subsection{Language Identification in the Trajectory Audit}

The trajectory audit identifies WebSearch query languages with the compressed
fastText \texttt{lid.176.ftz}
model~\citep{joulin2017bag,joulin2017fasttext}.  The released model recognizes
176 languages and was trained on Wikipedia, Tatoeba, and SETimes.  fastText
reports approximately 96.5\% accuracy on its labeled language-identification
evaluation.\footnote{\url{https://fasttext.cc/blog/2017/10/02/blog-post.html}}
For Wikipedia search results and fetched pages, the audit reads the language
directly from the hostname; path, bridge, and answer evidence is then detected
through the deterministic QID--URL--surface rules described in
Appendix~\ref{app:audit}.

\section{Complete Failure Breakdown}
\label{app:full-results}

Table~\ref{tab:full-failures} reports the 11 mutually exclusive leaves
for the original ten-model paper set as percentages.  All model-wise, prompt-wise, and pooled percentages
use the corresponding group's classifiable errors as denominator.
Table~\ref{tab:failure-mode} gives the exact percentages plotted in
Figure~\ref{fig:failure}.

\begin{table}[t]
\caption{Percentage composition of the 11 mutually exclusive failure leaves
by model.  Each percentage is conditioned on the classifiable errors in that
row.  Values may not sum to exactly
100\% because of rounding.}
\label{tab:full-failures}
\centering
\scriptsize
\setlength{\tabcolsep}{1.8pt}
\resizebox{\textwidth}{!}{%
\begin{tabular}{@{}lrrrrrrrrrrr@{}}
\toprule
\textbf{System} & \multicolumn{4}{c}{\textbf{Language covering (L)}}
& \multicolumn{5}{c}{\textbf{Evidence-path recovery (T)}}
& \multicolumn{2}{c}{\textbf{Residual (R)}} \\
\cmidrule(lr){2-5}\cmidrule(lr){6-10}\cmidrule(lr){11-12}
& \textbf{L1} & \textbf{L2} & \textbf{L3} & \textbf{L4}
& \textbf{T1a} & \textbf{T1b} & \textbf{T2a} & \textbf{T2b} & \textbf{T3}
& \textbf{R1} & \textbf{R2} \\
\midrule
Claude-Opus-4.8 & 1.2 & 14.3 & 24.8 & 40.4 & 0.6 & 9.3 & 6.2 & 2.5 & 0.6 & 0.0 & 0.0 \\
GLM-5.2 & 1.1 & 6.2 & 25.0 & 39.0 & 0.4 & 8.8 & 8.1 & 7.4 & 2.9 & 1.1 & 0.0 \\
GLM-5.3 & 0.8 & 5.0 & 20.9 & 41.8 & 0.4 & 10.9 & 9.6 & 7.5 & 2.5 & 0.0 & 0.4 \\
GLM-5.3-Flash & 0.3 & 4.1 & 20.7 & 40.5 & 2.0 & 11.2 & 9.9 & 8.2 & 2.4 & 0.3 & 0.3 \\
Qwen3.8-Max & 0.0 & 3.9 & 20.4 & 40.1 & 0.7 & 11.3 & 8.8 & 9.2 & 3.5 & 1.4 & 0.7 \\
DeepSeek-V4-Pro-0813 & 0.8 & 7.8 & 31.4 & 36.5 & 0.4 & 5.5 & 6.7 & 7.1 & 1.2 & 2.4 & 0.4 \\
DeepSeek-V4-Flash-0731 & 1.1 & 7.0 & 31.1 & 34.1 & 0.0 & 5.5 & 8.1 & 8.8 & 2.2 & 1.8 & 0.4 \\
GPT-5.6-Sol & 0.0 & 2.4 & 17.1 & 40.9 & 1.8 & 5.5 & 6.7 & 18.9 & 6.1 & 0.0 & 0.6 \\
Kimi-K3 & 0.0 & 1.2 & 9.3 & 27.9 & 0.4 & 7.3 & 14.6 & 22.7 & 9.7 & 1.2 & 5.7 \\
GPT-5.5 & 0.0 & 1.6 & 18.8 & 38.0 & 6.2 & 7.3 & 10.9 & 11.5 & 4.2 & 0.0 & 1.6 \\
\midrule
Pooled & 0.5 & 5.2 & 22.2 & 37.8 & 1.2 & 8.4 & 9.1 & 10.2 & 3.5 & 0.9 & 1.0 \\
\bottomrule
\end{tabular}
}
\end{table}

\paragraph{Subtype notation.}
L1--L4 denote source lock, English trap, wrong-language route, and incomplete
language cover.  T1a and T1b denote nominal switch
(\texttt{T1\_NOMINAL\_SWITCH}) and off-path retrieval
(\texttt{T1\_OFF\_PATH\_RETRIEVAL}).  T2a is a non-bridge path hit
(\texttt{T2\_NON\_BRIDGE\_PATH\_HIT}); T2b is partial branch coverage through
designated bridge entities
(\texttt{T2\_PARTIAL\_BRIDGE\_COVERAGE}).  T3 denotes answer-evidence failure.
R1 and R2 denote strong answer evidence in a fetched page
(\texttt{R\_STRONG\_FETCHED}) and only in search results
(\texttt{R\_STRONG\_SERP\_ONLY}), respectively.

\begin{table}[t]
\caption{Exact percentage composition shown in Figure~\ref{fig:failure}.
Every percentage is conditioned on the classifiable errors in that row.
As in the figure, T1 combines T1a and T1b, while T2 combines T2a and T2b.
Values may not sum to exactly 100\% because of rounding.}
\label{tab:failure-mode}
\centering
\scriptsize
\setlength{\tabcolsep}{2.5pt}
\resizebox{\textwidth}{!}{%
\begin{tabular}{@{}lrrrrrrrrr@{}}
\toprule
\textbf{Profile} & \textbf{L1} & \textbf{L2} &
\textbf{L3} & \textbf{L4} & \textbf{T1} & \textbf{T2} &
\textbf{T3} & \textbf{R1} & \textbf{R2} \\
\midrule
Claude-Opus-4.8 & 1.2 & 14.3 & 24.8 & 40.4 & 9.9 & 8.7 & 0.6 & 0.0 & 0.0 \\
GLM-5.2 & 1.1 & 6.2 & 25.0 & 39.0 & 9.2 & 15.4 & 2.9 & 1.1 & 0.0 \\
GLM-5.3 & 0.8 & 5.0 & 20.9 & 41.8 & 11.3 & 17.2 & 2.5 & 0.0 & 0.4 \\
GLM-5.3-Flash & 0.3 & 4.1 & 20.7 & 40.5 & 13.3 & 18.0 & 2.4 & 0.3 & 0.3 \\
Qwen3.8-Max & 0.0 & 3.9 & 20.4 & 40.1 & 12.0 & 18.0 & 3.5 & 1.4 & 0.7 \\
DeepSeek-V4-Pro-0813 & 0.8 & 7.8 & 31.4 & 36.5 & 5.9 & 13.7 & 1.2 & 2.4 & 0.4 \\
DeepSeek-V4-Flash-0731 & 1.1 & 7.0 & 31.1 & 34.1 & 5.5 & 16.8 & 2.2 & 1.8 & 0.4 \\
GPT-5.6-Sol & 0.0 & 2.4 & 17.1 & 40.9 & 7.3 & 25.6 & 6.1 & 0.0 & 0.6 \\
Kimi-K3 & 0.0 & 1.2 & 9.3 & 27.9 & 7.7 & 37.2 & 9.7 & 1.2 & 5.7 \\
GPT-5.5 & 0.0 & 1.6 & 18.8 & 38.0 & 13.5 & 22.4 & 4.2 & 0.0 & 1.6 \\
\midrule
Pooled & 0.5 & 5.2 & 22.2 & 37.8 & 9.6 & 19.3 & 3.5 & 0.9 & 1.0 \\
\texttt{ZH} & 0.9 & 7.0 & 16.7 & 51.3 & 6.0 & 13.4 & 2.7 & 0.7 & 1.3 \\
\texttt{EN} & 0.0 & 2.9 & 14.4 & 51.4 & 8.5 & 17.8 & 2.8 & 1.2 & 0.9 \\
\texttt{DL} & 0.8 & 6.1 & 34.3 & 13.5 & 13.5 & 25.3 & 4.8 & 0.8 & 0.8 \\
\bottomrule
\end{tabular}
}
\end{table}

\subsection{Correct-Trajectory Controls and Paired Process Analysis}
\label{app:correct-controls}

We apply the same path-aligned decision tree to every analyzable trajectory.
For a correct trajectory, the resulting leaf is stored as its behavioral
control label; for an incorrect trajectory, it is the reported failure leaf.
This common labeling rule supports direct comparisons between successful and
failed search processes.

\paragraph{T-stage enrichment is associated with failure.}
T-stage trajectories have a 54.4\% failure rate, compared with 39.2\% for
non-T trajectories, a difference of 15.2\,pp (95\% task-cluster bootstrap
CI: 8.5--21.8\,pp).  The marginal odds ratio is 1.85 (95\% CI:
1.41--2.44).  Equivalently, T accounts for 32.3\% of incorrect trajectories
and 20.5\% of correct controls.  The T share is higher among incorrect
trajectories for all ten models and under each of the \texttt{ZH},
\texttt{EN}, and \texttt{DL} prompt modes.  Incomplete post-route evidence
recovery is therefore consistently associated with failure across the
benchmark.  The largest leaf-level differences come from off-path retrieval
(T1b, $+5.7$\,pp) and non-bridge path entry (T2a, $+3.9$\,pp).  Together,
these two leaves account for 80.8\% of the total T-family enrichment.

\paragraph{Designated-language prompts advance the route and increase wrong-route entry.}
We pair each \texttt{DL} trajectory with the \texttt{ZH} and \texttt{EN}
trajectories for the same model and semantic task, retaining only pairs in
which both trajectories are analyzable.  This gives 1,680 \texttt{DL}--\texttt{ZH}
pairs and 1,848 \texttt{DL}--\texttt{EN} pairs spanning 199 semantic tasks.
Relative to \texttt{ZH},
\texttt{DL} increases route completion by 18.4\,pp, increases T-stage entry
by 17.0\,pp, and reduces incomplete-cover behavior by 35.8\,pp.  Relative to
\texttt{EN}, the corresponding changes are 15.3, 14.6, and
$-35.0$\,pp.  At the same time, wrong-language routing rises by 12.3\,pp
against \texttt{ZH} and 13.0\,pp against \texttt{EN}.  Every process shift
has the same direction for all ten models, and each task-cluster randomization
test gives $p<10^{-4}$.  The paired accuracy changes are much smaller:
$-1.7$\,pp against \texttt{ZH} and $+0.7$\,pp against \texttt{EN}.
Figure~\ref{fig:performance} reports prompt-mode accuracy over all 2,000
trajectories per mode (200 tasks $\times$ 10 models) under the benchmark
scoring policy.  The paired accuracy differences here retain only complete
process-analysis pairs: 1,680 \texttt{DL}--\texttt{ZH} pairs and 1,848
\texttt{DL}--\texttt{EN} pairs after removing any pair with an
\texttt{EXCLUDED} or \texttt{UNDETERMINED} member.  The two estimates
therefore use different denominators.
Thus, designated-language prompts move agents deeper into the search process
and also anchor more trajectories to an incorrect language route.

\begin{table}[t]
\caption{Paired effect of the designated-language prompt on process outcomes.
Each cell reports the \texttt{DL} percentage-point difference followed by its
95\% task-cluster bootstrap confidence interval.  Pairing uses the same model
and semantic task.  Appendix~\ref{app:sensitivity} gives the complete
resampling procedure.}
\label{tab:paired-dl-process}
\centering
\small
\setlength{\tabcolsep}{4.0pt}
\begin{tabular}{@{}lrr@{}}
\toprule
\textbf{Outcome} & \textbf{\texttt{DL} minus \texttt{ZH}} &
\textbf{\texttt{DL} minus \texttt{EN}} \\
\midrule
Route complete & $+18.4$ [14.2, 22.7] & $+15.3$ [11.0, 19.7] \\
T family & $+17.0$ [12.8, 21.2] & $+14.6$ [10.5, 18.9] \\
L3 wrong-language route & $+12.3$ [8.6, 16.2] & $+13.0$ [9.3, 16.9] \\
L4 incomplete cover & $-35.8$ [$-40.5$, $-31.3$] &
$-35.0$ [$-39.6$, $-30.4$] \\
Answer accuracy & $-1.7$ [$-4.3$, 1.0] & $+0.7$ [$-1.9$, 3.3] \\
\bottomrule
\end{tabular}
\end{table}

\paragraph{Success rises at successive post-cover gates.}
Among trajectories that complete a language route, entering the configured
evidence path raises empirical accuracy from 31.1\% to 56.6\%, a gain of
25.5\,pp.  Among trajectories that enter the path, complete bridge coverage
raises accuracy from 50.4\% to 70.0\%, a gain of 19.6\,pp.  Both associations
have the same direction for all ten models.  Among trajectories with complete
bridge coverage, strong answer exposure raises accuracy from 47.5\% to
83.1\%.  These transitions show that language-cover completion, path entry,
bridge recovery, and answer exposure contribute distinct stages of successful
search.

\begin{table}[t]
\caption{Conditional accuracy at successive post-cover gates.  The odds ratios
stratify by semantic task and prompt mode; confidence intervals use a
stratified bootstrap.}
\label{tab:post-cover-gates}
\centering
\small
\setlength{\tabcolsep}{3.5pt}
\resizebox{\textwidth}{!}{%
\begin{tabular}{@{}lrrrrr@{}}
\toprule
\textbf{Next gate} & \textbf{Passed (\%)} & \textbf{Not passed (\%)} &
\textbf{Difference (pp)} & \textbf{Stratified OR (95\% CI)} &
\textbf{Models with positive direction} \\
\midrule
Path entry after route completion & 56.6 & 31.1 & 25.5 &
1.78 [1.09, 3.05] & 10/10 \\
Complete bridges after path entry & 70.0 & 50.4 & 19.6 &
2.09 [1.24, 3.80] & 10/10 \\
Strong evidence after complete bridges & 83.1 & 47.5 & 35.6 &
2.71 [0.74, 12.57] & 9/10 \\
\bottomrule
\end{tabular}
}
\end{table}

\paragraph{Correct trajectories use route languages more directly.}
The median share of queries issued in required route languages is 25.0\% for
correct trajectories and 13.7\% for incorrect trajectories; the correct
group has the higher median for all ten models.  After entering a route
language, correct trajectories devote a smaller share of subsequent queries
to the prompt language (18.6\% versus 36.2\%) and to English (33.5\% versus
40.9\%).  These directions hold for nine of ten models.  Among trajectories
that enter a route language, the median first entry occurs at search action
four for correct trajectories and search action six for incorrect
trajectories.  Effective multilingual search therefore combines early route
entry, sustained route-language use, and limited fallback.

\paragraph{R provides a positive control for the strong-evidence gate.}
Across correct and incorrect trajectories, 83.1\% of R-labeled trajectories
produce the correct answer.  The label-conditional accuracy is 84.4\% for R1
fetched-page exposure and 81.7\% for R2 search-result-only exposure.  Strong
answer evidence is highly predictive of successful answer production through
both exposure channels.

\section{Classifier Implementation and Evidence Policy}
\label{app:audit}

The deterministic implementation parses WebSearch and WebFetch events,
identifies query languages, identifies fetched-page languages from URLs, and
maps explicit QIDs, Wikipedia URLs, and normalized multilingual surfaces to
the frozen task profile.

\paragraph{Route policy and observable language entry.}
Each profile specifies one or more task-level minimum language covers: the
smallest language sets that provide lexical access to all hidden retrieval-key
entities, as defined in Section~\ref{sec:generation}.  For trajectory auditing,
the final policy uses the same set as the \emph{runtime cover}.  A prompt
language contained in that set remains an explicit cover member and requires
an identifiable query or a fetched page whose URL identifies that language.
Prompt text is excluded from language entry.  Source-only retrieval is
classified as L1 for a non-English source and L2 for an English source.
Non-source route progress for L4 begins when retrieval enters an additional
required cover language.  The classifier uses the validated task-level covers
stored in the construction certificates.

For every configured runtime cover, the classifier records the first query or fetch
in each language.  The \emph{winning route} is the complete cover with the
earliest completion event (with deterministic tie breaking).  Route start is
the earliest language-entry event on that winning route, and route completion
is its latest required language-entry event.  Completion is the L/T gate, but
the path/bridge/answer evidence window opens at route start and continues to
the end of the trace.  This excludes unrelated earlier exposure while
retaining useful evidence acquired after route entry but before route
completion.

\paragraph{Entity and answer evidence.}
Path and bridge hits accept an explicit QID, a URL resolved to that QID, or a
normalized multilingual surface match.  Required bridge entities are designated
per evidence branch (graph \textsc{match}): a branch requirement is covered when
any designated bridge entity for that branch is hit, and full bridge coverage
requires every branch to be covered.
Strong answer evidence requires the event to contain the answer QID directly
or an exact normalized answer Wikipedia URL.  Alias-only answer mentions
remain auxiliary diagnostics and leave the T3/R strong-evidence gate unmet.

\paragraph{Language routing failures (L).}
Table~\ref{tab:ltr} summarizes the cascade; the implementation resolves the
overlap between L1 and L2 by assigning \textbf{L1 (source lock)} to
non-English runs whose identifiable retrieval remains in the source language
and \textbf{L2 (English trap)} to the corresponding English-source behavior.
L2 also covers English-prompt runs without non-source valid-route progress
unless they sustain a non-route-language branch, as well as non-English runs
that enter English without non-source valid-route progress or a sustained
non-route branch.
\textbf{L3 (wrong-language route)} applies when retrieval leaves the source but
makes no non-source progress toward a valid runtime cover and instead follows
non-route languages.
\textbf{L4 (incomplete language cover)} applies when the run enters at least
one required non-source language on a valid runtime cover but completes no
configured runtime cover.

\paragraph{Evidence-path failure gates (T).}
For trajectories that complete a route, \textbf{T1a (nominal switch)} applies when no
search-result or fetched-page content in a winning-route language appears in
the route-start window.  \textbf{T1b (off-path retrieval)} applies when such
content appears but no configured path entity is hit.
\textbf{T2a (non-bridge path hit)} applies when the path is entered but no
required evidence branch is covered by one of its designated bridge entities,
whereas \textbf{T2b (partial bridge coverage)} applies when some but not all
required branches are covered in this way.  \textbf{T3 (answer-evidence
failure)} requires a complete route, a path hit, and bridge-entity coverage of
every required branch, but no strong answer evidence.

\paragraph{Post-evidence residual failures (R).}
R contains incorrect runs that pass the complete-route, path, all-branches, and
strong-answer gates.  \textbf{R1 (fetched-page exposure)} applies if any strong
answer exposure in the route-start window is associated with a WebFetch event;
\textbf{R2 (search-result-only exposure)} applies when all strong exposures
occur in WebSearch results.  R is an observable-exposure label; model
understanding and evidence use lie outside its definition.

\paragraph{Exclusions and indeterminate cases.}
A missing exact trace or incomplete inference/semantic judgment is
\texttt{EXCLUDED}.  A profile with an unusable route or unusable bridge
annotation in its construction certificate is reported as
\texttt{UNDETERMINED}.

The reported failure analysis is defined over incorrect, classifiable runs and
assigns each one to its earliest failed gate.  Table~\ref{tab:ltr} summarizes
the branches, and
Appendix~\ref{app:sensitivity} reports supported sensitivity checks.

\section{Sensitivity Analyses}
\label{app:sensitivity}

\paragraph{Source-excluded versus source-included runtime routes.}
The final policy uses each task-level minimum language cover unchanged as the
runtime cover.  A prompt language contained in that cover requires an
observable query or page fetch.  The comparison policy treats the prompt
language as already satisfied and removes it from the runtime route.
Under both policies, source-only retrieval is classified as L1/L2.
Table~\ref{tab:source-policy}
compares these policies.  Both rows use the same original ten-model tasks and
the same classifiable-error set.

\begin{table}[t]
\caption{Ten-model classifiable-error composition under source-excluded and
source-included runtime-route policies.  L/T/R and L4
are percentages of the corresponding classifiable errors.}
\label{tab:source-policy}
\centering
\small
\setlength{\tabcolsep}{3.2pt}
\begin{tabular}{@{}lrrrr@{}}
\toprule
\textbf{Route policy} & \textbf{L (\%)} &
\textbf{T (\%)} & \textbf{R (\%)} & \textbf{L4 (\%)} \\
\midrule
Source excluded (comparison) & 65.06 & 32.97 & 1.97 & 37.09 \\
Source included (final) & 65.73 & 32.34 & 1.93 & 37.76 \\
\bottomrule
\end{tabular}
\end{table}

The headline family composition is stable under source retention, with every
family changing by less than 0.7 percentage points: L rises by 0.67 points, T
falls by 0.63, and R falls by 0.04.  L4 rises from 883 to 899 errors
(37.09\% to 37.76\%), reflecting 16 runs for which retrieval in a cover
language that is also the prompt language is required to complete the runtime
cover.

\paragraph{Statistical uncertainty and effective sample size.}
The semantic task is the primary experimental unit.  The benchmark contains
200 task clusters, each comprising three prompt renderings and outputs from
every system.  All uncertainty estimates resample these task clusters and
retain their dependent model--prompt trajectories together.

We construct 95\% confidence intervals with 10,000 task-cluster bootstrap
replicates.  Each replicate resamples the 200 semantic task IDs with
replacement and carries along all three prompt renderings and all model
outputs associated with each sampled task.  The same procedure is applied to
model accuracy, the highest--lowest model contrast, and L/T/R family shares.
We use random seed 20260924.

Model comparisons are paired because every system is evaluated on the same
semantic tasks.  We test accuracy differences using a
paired task-level randomization test with 10,000 sign-flip replicates and
report effects in percentage points.  The reported $p$-value is Holm-adjusted
over all 45 model pairs.

\paragraph{Correct-control and paired-process analyses.}
Appendix~\ref{app:correct-controls} defines analyzable trajectories by
classifiable primary labels; \texttt{EXCLUDED} and \texttt{UNDETERMINED}
remain outside this set.  Correct trajectories retain
\texttt{CORRECT\_CONTROL} as their primary label and use the behavioral leaf
produced by the same L/T/R tree used for incorrect trajectories.  The marginal
T-stage failure-rate difference and odds ratio use all such trajectories.
Their 95\% intervals come from 10,000 bootstrap replicates over the analyzable
semantic-task clusters, carrying all available model and prompt-mode
trajectories with each sampled task.

The \texttt{DL} comparisons pair analyzable runs from the same model and
semantic task.  This produces 1,680 \texttt{DL}--\texttt{ZH} pairs and 1,848
\texttt{DL}--\texttt{EN} pairs across the analyzable semantic tasks.  Their confidence
intervals use 10,000 task-cluster bootstrap replicates.  The two-sided
randomization tests first aggregate paired differences within each semantic
task and then independently sign-flip the task-level differences for
10,000 replicates.  Figure~\ref{fig:performance} reports answer accuracy over
the cases retained by the scoring policy in each model--prompt cell.  Because
unrecovered errors other than timeouts and refusals are skipped under the
exception policy in Appendix~\ref{app:agent-setup}, a cell's denominator may
be smaller than 200.  The paired accuracy row in
Table~\ref{tab:paired-dl-process} instead uses the 1,680
\texttt{DL}--\texttt{ZH} and 1,848 \texttt{DL}--\texttt{EN} complete
process-analysis pairs defined above.

For Table~\ref{tab:post-cover-gates}, each row retains trajectories that pass
the preceding gate.  We form semantic-task--prompt-mode strata and report the
Mantel--Haenszel common odds ratio for passing the next gate versus answer
correctness.  Its percentile interval uses 10,000 bootstrap resamples of
these strata.  All correct-control and paired-process resampling uses random
seed 20260924.

\begin{table}[t]
\caption{Task-cluster uncertainty and paired comparisons.  Confidence
intervals resample semantic tasks while retaining all dependent prompt
renderings and model outputs.  Reported $p$-values are Holm-adjusted.}
\label{tab:statistical-uncertainty}
\centering
\scriptsize
\setlength{\tabcolsep}{4.0pt}
\begin{tabular}{@{}lrrr@{}}
\toprule
\textbf{Quantity or paired contrast} & \textbf{Estimate} &
\textbf{95\% CI} & \textbf{Adjusted $p$} \\
\midrule
Claude-Opus-4.8 accuracy & 69.7 & [64.0, 75.2] & -- \\
DeepSeek-V4-Flash accuracy & 33.8 & [28.2, 39.5] & -- \\
Claude-Opus-4.8 $-$ DeepSeek-V4-Flash & 35.8\,pp &
[30.0, 41.5] & 0.004 \\
Language-covering failures (L) & 65.7 &
[60.4, 70.9] & -- \\
Evidence-path failures (T) & 32.3 &
[27.2, 37.6] & -- \\
Residual failures (R) & 1.9 &
[1.1, 3.0] & -- \\
\bottomrule
\end{tabular}
\end{table}

After Holm correction, the 35.8-point highest--lowest model gap is significant.
The lower confidence bound for L remains above the upper bound for T, and the
upper confidence bound for R remains below 3\%.

\section{Failure-Mode and Case Studies}
\label{app:cases}

This appendix grounds the audit in individual trajectories.  We first map all
audit boundaries onto one certified evidence path, then give one observed case
for every failure type in Table~\ref{tab:ltr}, and finally give
matched-prompt cases for the main findings.  Each case reports its queries,
retrieved pages, entity hits, termination state, and score.  Instance
identifiers refer to the shared semantic task; prompt modes identify its
\texttt{ZH}, \texttt{EN}, or \texttt{DL} rendering.

\subsection{Reading the Audit on One Certified Evidence Path}
\label{app:ltr-walkthrough}

\paragraph{Task and certificate.}
Consider the following simplified, graph-aligned rendering of
\texttt{rare100-graph041}:
\begin{quote}
\emph{A church named after Archangel Gabriel is located in a town.  A person
who holds the position of hieromonk was born in a village located within the
administrative territory of the same town.  In which country is that town
located?}
\end{quote}
The answer is \emph{Bosnia and Herzegovina} (\texttt{Q225}).  The first branch
contains Saint-Gabriel Church (\texttt{Q19950990}), which is named after
Gabriel (\texttt{P138}) and located in \v{C}elinac
(\texttt{Q338457}, \texttt{P131}).  The second contains Iguman Antonije
(\texttt{Q135973749}), whose position is hieromonk (\texttt{P39}) and whose
birthplace is Grabovac (\texttt{Q3112965}, \texttt{P19}); Grabovac is also
located in \v{C}elinac (\texttt{P131}).  The two branches therefore converge
on the same hidden town, whose country relation (\texttt{P17}) yields
\texttt{Q225}.

The graph entities and relations are language-neutral.  Language dependence
comes from the frozen lexical interface.  The certified task-level minimum
language cover is French plus Serbian: the catalogued surface for the church is French
(\emph{\'eglise Saint-Gabriel de \v{C}elinac}), while the catalogued surface
for Antonije is Serbian (\emph{Iguman Antonije} in Serbian script).  They are
the two required bridge entities, one per evidence branch.
Here, a ``bridge'' is the certificate-designated entity connecting retrieval
to an evidence branch.  The trajectory audit uses this same set as the runtime
cover; if French is also the prompt language, an
observed French query or page fetch is still required to enter it.

\paragraph{Where the same task can fail.}
The certificate induces the following failure cascade:
\begin{itemize}
    \item \textbf{No non-source valid-route progress (L1/L2).}  Retrieval remains at the
    source-language boundary.  A non-English source lock is L1; an English
    source lock or English fallback with no required non-source route-language
    entry and no sustained non-route-language branch is L2.
    An English-prompt run that never leaves English is therefore L2, not L1.
    \item \textbf{Wrong route (L3).}  The run leaves the source environment
    but sustains retrieval in a language outside every valid runtime cover, without
    entering French or Serbian.
    \item \textbf{Incomplete valid cover (L4).}  The run enters part of the
    certified route beyond the prompt language---for example, under the French
    \texttt{DL} rendering, Serbian but not French through an observed retrieval
    action---yet never completes the runtime cover.
    \item \textbf{Off-path retrieval after complete cover (T1).}  French and
    Serbian have both been entered, but the retrieved results and pages expose
    no configured path entity.
    \item \textbf{Incomplete bridge recovery (T2).}  The trace reaches a path
    entity such as Gabriel, hieromonk, or Grabovac, but does not cover both
    required evidence branches through their designated bridge entities.
    This includes coverage of zero or one of
    \texttt{Q19950990} and \texttt{Q135973749}.
    \item \textbf{Answer-evidence failure (T3).}  Both required evidence
    branches are covered through their designated bridge entities, but neither
    answer QID \texttt{Q225} nor an exact answer Wikipedia URL appears.
    \item \textbf{Residual failure after answer exposure (R).}  The complete
    runtime cover, both required evidence branches through their designated
    bridge entities, and strong answer evidence are all observed, yet the run
    receives score zero.  R1 places the strong evidence in a fetched page; R2
    leaves it only in search results.
\end{itemize}
The classifier assigns the earliest unsatisfied boundary in this cascade.
Thus the case separates whether the agent completes a valid runtime cover,
reaches the configured path, covers every required evidence branch through a
designated bridge entity, sees strong answer evidence, and converts that
evidence into a correct completion.

\subsection{A Cross-Language Web-Search Case}
\label{app:web-scope-stress}

\paragraph{Task and evidence route.}
The English rendering reads:
\begin{quote}
\itshape
An official who had served in the judicial institutions of an autonomous
region was sentenced to corrective labor during a political purge and later
died in a large labor camp.  A veterinarian who had served as a municipal
head also died in the same camp system; records about the two are preserved
in archives in different places.

The camp initially used an industrial mining settlement as its administrative
base, then moved its administration along the camp railway to a newly built
settlement.  What is the present-day name of the settlement that served as its
long-term administrative center?
\end{quote}

Russian records
identify Matvey Spiridonov, a former judicial official who died in Vyatlag;
Latvian records identify Teodors Ermansons, a veterinarian and former
J\=ekabpils mayor who died in the same camp system.\footnote{\href{https://www.jgb.lv/lv/novadpetnieciba-kalendars/ieraksts/ermansons-teodors/}{Latvian local-history record for Teodors Ermansons.}}
The question then asks for the modern name of the settlement to which the camp
administration moved from Rudnichny along its railway.  The answer is
\emph{Lesnoy}.  The frozen
certificate assigns the two person branches to a Latvian--Russian minimum
cover; removing either graph branch leaves at least 14 candidate answers.

\paragraph{Search verification.}
We replay the clues as separate Chinese, English, Latvian, and Russian queries
and inspect the search results, matching the
information exposed by \texttt{WebSearch}.  The
Chinese and English queries return other Gulag systems.  The initial Russian
query also fails to identify the modern settlement.  The Latvian query
\emph{Teodors Ermansons veterin\=ar\=arsts pils\=etas galva miris Vjatlag\=a}
does identify Ermansons and reveals that he died in \emph{Vjatlags}, but its
results do not contain the requested modern name \emph{Lesnoy}.

This missing camp name supplies the next query in Russian.  Searching for the
movement of the Vyatlag administration from Rudnichny along the camp railway
returns result snippets stating that the administration moved to
Lesnaya/Sotsgorodok and that the settlement was renamed \emph{Lesnoy} in
1944.\footnote{\href{https://www.abramichi.ru/raznoe/veremjev_vjatlag.php}{Russian Vyatlag history returned by the follow-up query.}}
The answer therefore first appears only after a Latvian-to-Russian transition:
Latvian search identifies the shared camp, while Russian search supplies the
camp-history fact needed to obtain its modern administrative center.

\paragraph{Observed trajectories.}
The model traces show the same pattern.  Only 9 of the 30 model--prompt runs
answer correctly.  The remaining runs fail at several points:
some never complete the language cover, while others reach the path but miss
bridge or answer evidence.  In the fixed search-result audit, the answer is
exposed only after the Latvian-to-Russian transition; this case therefore
requires cross-language search.

\subsection{One Case for Each Failure Type}
\label{app:subtype-cases}

\subsubsection{Language Routing}

\paragraph{L1: Source lock.}
For \texttt{main100-graph081}, GLM-5.2 receives the Czech \texttt{DL}
rendering of a pear-identification task.  The admissible covers are
Czech--French and Czech--Hungarian.  All 65 classifier-identifiable queries
and all three classified page fetches are Czech; neither French nor Hungarian
is entered.  The run has score zero and terminates \texttt{finished}.  This is
L1 because all retrieval remains in the non-English source language.

\paragraph{L2: English trap.}
For \texttt{main100-graph011}, Claude-Opus-4.8 receives the English rendering
of a gallery-identification task whose task-level minimum cover---Persian,
Russian, Turkish, and Ukrainian---also serves as its runtime cover.  All 439
identifiable queries and the sole classified
page fetch are English.  No required non-source route language and no sustained wrong
language is entered.  The score-zero run terminates \texttt{finished}; its
reference is \emph{Akanat Gallery}.  It is therefore an English-source lock,
L2.

\paragraph{L3: Wrong-language route.}
On \texttt{main100-graph050}, Claude-Opus-4.8 receives a Chinese prompt whose
two clue chains identify a river through Hungarian and Slovak evidence.  After
brief Chinese and English searches, the run follows a sustained Russian
branch: it issues 15 Russian queries and fetches three Russian pages about
North Caucasus museums, but never queries in Hungarian or Slovak.  It
has score zero and terminates \texttt{finished}; the reference is the
\emph{Horn\'ad} (\emph{Hern\'ad}).  With no Hungarian or Slovak entry, the
sustained Russian branch places the case at L3.

\paragraph{L4: Incomplete valid cover.}
For \texttt{rare100-graph074}, Claude-Opus-4.8 receives an English prompt for
a task whose sole task-level minimum cover, Dutch plus Ukrainian, also serves
as its runtime cover.  The run switches to
Ukrainian at its second event and accumulates 36 Ukrainian queries and 31
Ukrainian page fetches, but it never enters Dutch.  One substantial
target-language branch is therefore still only a partial cover.  The run has
score zero, terminates \texttt{finished}, and is L4.

\subsubsection{Evidence-Path Recovery after Complete Language Cover}

\paragraph{T1: Off-path retrieval.}
On \texttt{main100-graph066}, Claude-Opus-4.8 receives the German
\texttt{DL} rendering.  German at the start and a Serbian query at sequence
179 complete the German--Serbian route.  The winning-route window contains 17
route-language content events, but no configured path QID appears.  The run
has score zero against \emph{Syrius Eberle} and terminates
\texttt{finished}.  It therefore has both a complete cover and route-language
content, but the content remains off the certified path: T1.

\paragraph{T2: Incomplete bridge recovery.}
On \texttt{rare100-graph073}, Claude-Opus-4.8 receives the French
\texttt{DL} rendering and completes the French--Ukrainian cover.  Ukrainian
retrieval exposes the Schmalhausen Institute of Zoology and the ornithologist
Anatolii Poluda (\texttt{Q30958268}), covering one required evidence branch
through its designated bridge entity.  The other required branch, whose
designated bridge entity is \texttt{Q98609033}, is never covered.  The run
terminates \texttt{finished} with score zero against the reference
\emph{Zoological-Biological Institute of the All-Ukrainian Academy of
Sciences}.  Reaching the path while leaving one of two required evidence
branches uncovered is T2.

\paragraph{T3: Complete bridges without strong answer evidence.}
For \texttt{rare100-graph027}, Claude-Opus-4.8 receives the Catalan
\texttt{DL} rendering and completes the Catalan--Spanish cover.  Spanish
retrieval hits Javier Tr\'{\i}mboli on one branch and Plaza Alfredo Nobel or
Homenaje a la Bandera on the other, covering both required branches through
designated bridge entities.
Neither the answer QID nor an exact answer Wikipedia URL for
\emph{In\'es Fern\'andez Moreno} appears.  The run has score zero, terminates
\texttt{finished}, and is therefore T3.

\subsubsection{Residual Failure after Strong Evidence Exposure}

\paragraph{R1: Strong fetched-page exposure.}
\texttt{rare100-graph042} from GLM-5.3-Flash under the Chinese prompt
completes its Spanish--Hebrew minimum language cover, covers both required
evidence branches,
and at sequence 62 fetches the exact Spanish Wikipedia page for
\emph{Villa Luro}.  The run receives score zero and terminates
\texttt{inference\_timeout}.  Strong evidence on an opened page makes this R1.

\paragraph{R2: Strong search-result-only exposure.}
For \texttt{rare100-graph090}, Kimi-K3 receives the Lithuanian
\texttt{DL} rendering and completes the Lithuanian--Russian cover.  The trace
eventually covers both required evidence branches through the designated
bridge entities \texttt{Q4272834} and
\texttt{Q9273854}.  A search result at sequence 910 exposes the profile's
exact Russian Wikipedia URL for \emph{Svir Castle Hill}, but the answer page
is never fetched.  The score-zero run terminates \texttt{finished}; only the
search-result channel contains strong answer evidence, so this is R2.

\subsection{Matched-Prompt Cases for the Main Claims}
\label{app:claim-cases}

The preceding cases explain individual failure types.  We next compare
parallel renderings of the same semantic task to show how prompt language
changes the Web surface exposed during search and where the trajectory stops
in the failure cascade.

\paragraph{Complete language cover need not complete the evidence chain.}
For \texttt{rare100-graph023}, we compare GPT-5.5's English and Chinese
renderings of the same \emph{Lower Dniester National Park} task.  Both traces
complete the common Romanian--Ukrainian minimum language cover.  The English trace then has only one
route-language content event and no configured path hit, so its final label is
T1.  The Chinese trace reaches three path entities and covers one of two
required evidence branches through a designated bridge entity, so it is T2,
not R, even though strong search-result evidence also
appears elsewhere in the trace.  Both have score zero and record
\texttt{inference\_timeout}.  This matched case shows why language
cover, path entry, complete branch coverage, and answer exposure remain
separate audit boundaries.

\paragraph{One language switch is not a complete cover.}
On \texttt{rare100-graph081}, Qwen3.8-Max searches for the Radun (Radin)
Yeshiva under the required Russian--Turkish cover.  The Turkish \texttt{DL}
run has five English queries, two English page fetches, and only one Turkish
query; it never enters Russian and ends \texttt{finished} with score zero at
L2: the Turkish query enters only the source member of the cover, while no
required non-source route language is entered.  The Chinese run does switch
into Russian, but never Turkish, so it ends
\texttt{finished} with score zero at L4.  These matched traces show that one
multilingual pivot does not complete the certified task-level two-language
cover.

\paragraph{Designated-language prompts can shift L4 failures to L3.}
On \texttt{rare100-graph058}, Claude-Opus-4.8 searches for the
\emph{Church of the Sacred Heart of Jesus in Krasyliv} under a
Dutch--Polish minimum language cover.  The English and Chinese runs issue 22
and four Polish queries, respectively, but never enter Dutch.  Both runs
terminate \texttt{finished} with score zero at L4.  Under the Polish
\texttt{DL} rendering, the trace again never enters Dutch and instead
sustains a Ukrainian--Russian branch, issuing 47 Ukrainian and 16 Russian
queries about the town, church, and aviation clues.  This score-zero,
\texttt{finished} run ends at L3.  The matched trio shows how a designated
language shifts the observed failure profile from L4 to L3: the trace starts
from one cover language but anchors subsequent search to plausible non-cover
languages without reaching the missing Dutch branch.

\section{Detailed Prompts}
\label{app:prompts}

\begin{promptbox}
You are a highly capable search agent that handles research tasks
of any depth or breadth---from deep multi-step investigations to
broad cross-entity surveys. End with a clear, calibrated answer
grounded in the evidence you actually used.

\medskip
The current date is \PromptCode{\{today\_date\}}.

\PromptHeading{\# Your strengths}

\begin{itemize}[
  leftmargin=1.4em,
  itemsep=2pt,
  topsep=2pt,
  parsep=0pt
]
  \item Breaking complex questions into a tractable line of
  investigation, or a flat set of parallel sub-investigations
  when scope is wide.

  \item Following leads through to primary sources, and pulling
  the specific facts out of long pages.

  \item Synthesizing across multiple sources when they disagree
  or no single one is authoritative, weighing credibility and
  recency to reconcile findings.

  \item Producing a calibrated final answer that distinguishes
  what is well-supported from what remains uncertain.
\end{itemize}

\PromptHeading{\# Doing research tasks}

\begin{itemize}[
  leftmargin=1.4em,
  itemsep=2pt,
  topsep=2pt,
  parsep=0pt
]
  \item Keep each investigation anchored to the user's original
  question.

  \item Think carefully through the task, and you MUST use tools
  to help your reasoning before solving it.

  \item When in doubt, search; confidence is not a substitute for
  verification. Do not answer from memory; use retrieved evidence
  as the source of truth for factual claims, candidate answers,
  entity names, dates, numbers, rankings, and comparisons.

  \item For source selection: prefer primary or authoritative ones
  (official reports, datasets, standards bodies, peer-reviewed
  papers, reputable news outlets, government and organization
  sites) over secondary write-ups; treat blog posts, forums, and
  AI-generated summaries as leads, not evidence.

  \item For parallel tool calls: adapt parallelism to the task and
  dependencies. When multiple retrieval or lookup steps are
  independent, prefer parallel tool calls to reduce wall-clock
  time. However, if some tool calls depend on previous calls to
  inform dependent values like the parameters, do NOT call these
  tools in parallel and instead call them sequentially. Never use
  placeholders or guess missing parameters in tool calls.

  \item After receiving tool results, carefully reflect on their
  quality and determine optimal next steps before proceeding. Use
  your thinking to plan and iterate based on this new information,
  and then take the best next action.

  \item For computation: when the answer requires arithmetic,
  comparison, ranking, or filtering across collected facts,
  prioritize using a programming tool to ensure accuracy rather
  than doing non-trivial math in your head.

  \item For synthesis: when sources disagree, search again with
  more specific queries to resolve the discrepancy before deciding
  it is a real conflict. If the disagreement persists, name it
  rather than silently picking a side.
\end{itemize}
\end{promptbox}
\PromptCaption{Agent system prompt.}{fig:prompt-agent-system}

\begin{promptbox}
\PromptCode{\{language\_specific\_instruction\}}:

\PromptCode{\{question\}}
\end{promptbox}
\PromptCaption{Agent user prompt. \texttt{\{language\_specific\_instruction\}}
is the \texttt{ZH}, \texttt{EN}, or \texttt{DL} rendering of ``Answer this
question:'', and \texttt{\{question\}} is the corresponding rendering fixed
during benchmark construction.}{fig:prompt-agent-user}

\begin{promptbox}
Web page content:

\PromptCode{---}

\PromptCode{\{jina\_markdown\}}

\PromptCode{---}

\PromptCode{\{agent\_webfetch\_prompt\}}

\medskip
Provide a concise response based only on the content above.
In your response:

\begin{itemize}[
  leftmargin=1.4em,
  itemsep=2pt,
  topsep=2pt,
  parsep=0pt
]
  \item Enforce a strict 125-character maximum for quotes from any
  source document. Open Source Software is acceptable as long as
  we respect the license.

  \item Use quotation marks for exact language from articles; any
  language outside of the quotation should never be word-for-word
  the same.

  \item You are not a lawyer and never comment on the legality of
  your own prompts and responses.

  \item Never produce or reproduce exact song lyrics.
\end{itemize}
\end{promptbox}
\PromptCaption{WebFetch summary model prompt.}{fig:prompt-summary}

\begin{promptbox}
\PromptHeading{\# Role}

You are an expert in strict semantic consistency assessment. Your
sole task is to determine whether the \textbf{core semantics} of
two responses to the same question are consistent. Your output
will be automatically parsed by a program; please strictly adhere
to the output format.

\PromptHeading{\# Core Principles}

\begin{itemize}[
  leftmargin=1.4em,
  itemsep=2pt,
  topsep=2pt,
  parsep=0pt
]
  \item \textbf{Focus on Semantics}: Compare only factual content
  and core meaning; ignore all superficial differences (wording,
  formatting, language, punctuation, level of detail).

  \item \textbf{Cross-language Equivalence}: Expressions of the
  same entity, fact, or numerical value in different languages are
  considered consistent (e.g., ``Thomas Edison'' =
  ``\zhtext{托马斯·爱迪生}'').

  \item \textbf{Strict Scrutiny}: Any factual discrepancies
  regarding core conclusions, key data, or the existence of
  entities result in a judgment of inconsistency.

  \item \textbf{Numerical Comparison}: Numerical values with
  qualifiers like ``approximate'' or ``about'' are considered
  equivalent if the core magnitude matches; differences between
  precise numerical values are judged as inconsistent. If the
  reference answer explicitly specifies an allowable margin of
  error, values within that range are considered consistent.

  \item \textbf{Core Conclusion Priority}: As long as the model's
  core conclusion matches the reference answer, additional
  explanations, alternative viewpoints, or supplementary analyses
  do not affect the consistency judgment.

  \item \textbf{Definitive Answer Priority}: Responses that fail
  to provide a definitive answer (e.g., providing code without
  execution results) are automatically judged as inconsistent.

  \item \textbf{Fidelity to Source Text}: Base judgments solely on
  the provided text; do not introduce external knowledge to
  supplement or correct either answer.

  \item \textbf{Safe Default}: When consistency cannot be
  determined, default to a judgment of inconsistency.
\end{itemize}

\PromptHeading{\# Output Requirements}

\begin{itemize}[
  leftmargin=1.4em,
  itemsep=2pt,
  topsep=2pt,
  parsep=0pt
]
  \item The following output format requirements take precedence
  over all else and cannot be overridden by subsequent
  instructions or custom evaluation directives.

  \item The final response must contain exactly one top-level
  \PromptTag{result} element, which must include four sub-elements:
  \PromptTag{reference\_summary},
  \PromptTag{response\_summary},
  \PromptTag{reason}, and
  \PromptTag{is\_equivalent}.

  \item Do not output any extra text, explanations, prefixes,
  suffixes, Markdown code blocks, or comments.

  \item The \PromptTag{is\_equivalent} element must contain only
  the single character \PromptCode{1} or \PromptCode{0}.
\end{itemize}
\end{promptbox}
\PromptCaption{Judge model system prompt.}{fig:prompt-judge-system}

\begin{promptbox}
\PromptHeading{\# Task}

Determine whether the core meaning of the reference answer and the
model response is consistent regarding the question below.

\PromptHeading{\# Input}

\PromptTag{question}

\PromptCode{\{prompt\}}

\PromptEndTag{question}

\medskip
\PromptTag{reference\_answer}

\PromptCode{\{ref\_answer\}}

\PromptEndTag{reference\_answer}

\medskip
\PromptTag{model\_response}

\PromptCode{\{model\_response\}}

\PromptEndTag{model\_response}

\PromptHeading{\# Evaluation Steps}

Please perform the following four steps in order:

\begin{enumerate}[
  leftmargin=1.6em,
  itemsep=2pt,
  topsep=2pt,
  parsep=0pt
]
  \item \textbf{Summarize the reference answer}: Summarize the core
  conclusion of the reference answer regarding the question into
  a concise summary (25 characters or fewer). Base the summary
  solely on the original text; do not introduce external knowledge.
  If the reference answer does not provide a clear answer, the
  summary should be ``No clear answer.''

  \item \textbf{Summarize the model response}: Process the model
  response in the same way. If the model response provides only
  code without explicitly stating the execution result, the
  summary should be ``No clear answer.''

  \item \textbf{Determine consistency}: Compare the core meanings
  of the two summaries and output the result based on the criteria
  below.

  \item \textbf{Self-check}: Review whether the two summaries
  faithfully reflect the original texts. If discrepancies are
  found, correct them before outputting the final result.
\end{enumerate}

\PromptHeading{\# Consistency Criteria}

\begin{itemize}[
  leftmargin=1.4em,
  itemsep=2pt,
  topsep=2pt,
  parsep=0pt
]
  \item Core meanings are completely equivalent
  $\rightarrow$ Consistent.

  \item Different forms of expression
  (language/format/level of detail) but identical meaning
  $\rightarrow$ Consistent.

  \item Model response's core conclusion is correct, with
  additional alternative viewpoints, explanations of controversy,
  or supplementary analysis $\rightarrow$ Consistent.

  \item Reference answer explicitly states a margin of error, and
  the model response's value falls within that range
  $\rightarrow$ Consistent.

  \item Reference answer does not state a margin of error, and
  there is a numerical difference (even if small)
  $\rightarrow$ Inconsistent.

  \item Discrepancies in key data or conclusions
  $\rightarrow$ Inconsistent.

  \item Reference answer contains multiple points, but the model
  response covers only some of them $\rightarrow$ Inconsistent.

  \item Either side is ``No clear answer''
  $\rightarrow$ Inconsistent.

  \item Unable to determine $\rightarrow$ Inconsistent.
\end{itemize}

\PromptHeading{\# Examples}

\noindent\textbf{\#\# Example 1: Cross-language + Unit conversion
$\rightarrow$ Consistent}

\medskip
\PromptTag{question}What is the speed of light?\PromptEndTag{question}

\medskip
\PromptTag{reference\_answer}The speed of light in a vacuum is
approximately 300,000 kilometers per second.\PromptEndTag{reference\_answer}

\medskip
\PromptTag{model\_response}The speed of light is approximately
$3\times10^{8}$ m/s.\PromptEndTag{model\_response}

\medskip
\PromptTag{result}

\PromptTag{reference\_summary}The speed of light is approximately
300,000 km/s.\PromptEndTag{reference\_summary}

\PromptTag{response\_summary}The speed of light is approximately
$3\times10^{8}$ m/s.\PromptEndTag{response\_summary}

\PromptTag{reason}The numerical value is the same; only the units
and language differ.\PromptEndTag{reason}

\PromptTag{is\_equivalent}1\PromptEndTag{is\_equivalent}

\PromptEndTag{result}

\medskip
\PromptCode{[10 more examples hidden for brevity]}
\end{promptbox}
\PromptCaption{Judge model user prompt.  The evaluation prompt includes 11
examples covering typical judging situations; one is shown here and ten are
elided for brevity.}{fig:prompt-judge-user}

\end{document}